Universität Rostock
Traditio et Innovatio

# Master Thesis

## Exploring transfer learning for Deep NLP systems on rarely annotated languages

Submitted by:
**Dipendra Yadav**

Submitted on:
20.07.2020

Supervisors:
Dr. Tobias Strauß
Dr.-Ing. Kristina Yordanova

University of Rostock | Department of Computer Science and Electrical Engineering
Institute of Visual & Analytic Computing | Mobile Multimedia Information Systems
Albert-Einstein-Str. 22 18059 Rostock | Tel: +49 381 498 7511

# Acknowledgements

I would like to thank following people for their help in different ways for the implementation of this Master Thesis:

**Dr. -Ing. Kristina Yordanova**
Dr. Yordanova acted as the supervisor for this thesis work at the University of Rostock. I am grateful for her guidance and support throughout the work in the form of meetings, technical advice and review of the implementation of the work.

**Dr. Tobias Strauß**
Dr. Strauß acted as the supervisor of this thesis work at PLANET AI. I am very grateful to him for providing me the initial guidance to explore different methods for the implementation of this work and helping me in getting started with it by providing relevant literature references. I would also like to thank him for especially helping me in determining the scope of the work and providing necessary suggestions with the required technical modifications. I would also like to thank the whole PlanetAI family for the warmth and support provided to me throughout my work.

I would like to dedicate this work to my family and friends.

Rostock, 20.07.2020
Dipendra Yadav

# Abstract

**Natural language processing** has seen exponential growth since deep learning has been used for various tasks. Deep learning has replaced the old rule-based methods by providing better performance on almost all the tasks in NLP, as deep learning basically looks into the hidden patterns and structures in the data which were sometimes not detected and at other-times overlooked by the rule-based systems. The research and development in NLP have been however focused on some handful languages of the world given their financial importance and a huge number of speakers. This has left out most of the other languages unexplored. The other reason for the lack of research in these languages is the unavailability of properly annotated and the required amount of datasets which is very important for training deep learning neural networks. The situation looks grim. When we look at the world language tree, thankfully we see that most of the unexplored languages have other well-explored languages nearby which basically means that both the languages are kind of similar. So if we could somehow adapt research and development from one language to another, it could be really helpful for both the languages. The goal of this thesis work is to explore transfer learning for the task of Part-of-speech(POS) tagging between Hindi and Nepali languages which are part of the Indo-Aryan language family with very high similarity. We have tried to explore the possibility of jointly training a model for the task of POS tagging in both Hindi and Nepali language and see if it helps in improving the performance of the model. We also try to explore if multitask learning in the Hindi language can be helpful for the task of POS tagging with auxiliary tasks of the gender tagging and singularity/plurality tagging. The deep learning architecture used for this work is BLSTM-CNN-CRF. The model is trained with monolingual word embeddings, vector mapped embeddings, and also with jointly trained Hindi-Nepali word embeddings in different setups with varying dropout from 0.25 to 0.5 and using ADAM and AdaDelta optimizers. These modes were used to train the models of individual languages, multitask learning in the Hindi language, and transfer learning for the task of POS tagging between Hindi and Nepali language. It has been observed that the jointly trained Hindi-Nepali word embeddings seem to improve the performance of all the models, better than monolingual word embeddings and vector mapped word embeddings. The default setup derived from Reimers et. al. [1] seems to help the model in its task of POS tagging. It has also been observed that multitask learning doesn't seem to help in the task of POS tagging in the Hindi language, on the contrary, it degrades the performance. And for jointly training a model for the task of POS tagging between Hindi and Nepali language, it also doesn't seem to help in the performance of the model.

# Contents

# List of Abbreviations

| | |
|---|---|
| AI | Artificial Intelligence |
| NLP | Natural Language Processing |
| POS | Part Of Speech |
| NER | Named Entity Recognition |
| LSTM | Long-Short Term Memory |
| Bi-LSTM | Bi-directional Long-Short Term Memory |
| CNN | Convolutional Neural Network |
| RNN | Recurrent Neural Network |
| CRF | Conditional Random Field |
| SGD | Stochastic Gradient Descent |
| ADAM | Adaptive Moment Estimation |
| HMM | Hidden Markov Model |
| MTL | Multitask Learning |
| CBOW | Continous Bag Of Word |
| LDC | Linguistic Data Consortium |
| CLE | Center for Language Engineering |
| TL | Transfer Learning |

# List of Figures

# List of Tables

# 1 Introduction

## 1.1 Overview

Human beings are considered one of the most intelligent species on the planet Earth. The ability to communicate complex ideas using natural languages is considered the trademark of human intelligence. Hence the ability to work with human languages is considered one of the most important features in the field of artificial intelligence(AI), which is concerned with building intelligent machines that can augment human intelligence. The field of AI which is concerned with the above task of programming computers to handle and interpret a large amount of natural language texts is called natural language processing(NLP) [2].

Given the diversity of natural languages in the world, it is often very difficult even for humans to share their knowledge among each other if they don't speak the same language. When it comes to intelligent machines, the task gets even tougher due to the ambiguity. But there are languages which are close in the language family tree which uses similar words and script. Hindi and Nepali are one of the pair which has these characteristics and it is interesting to explore if technological progress in one could be transferred to the other.

## 1.2 Motivation

The human language is incredibly diverse. As if in the year 2020, there are around 7,117 languages spoken around the world [3]. However, linguistic research has been essentially focused on the most predominantly used languages like English, German, Spanish, Chinese, etc. given their financial values and the total number of speakers [3]. This has led to the great language divide[1] which can be seen most evidently online where only a handful of languages are used by most of the people in the world for communication. As the other languages don't even have enough available dataset to conduct proper research.
Hence, this thesis work tries to explore if it is possible for a deep neural network architecture trained for one particular task in one language to perform better on the same task in the different languages given that the languages exhibit similarities. This can help if one language has fewer resources compared to the other language. This research work tries to find out if it is possible to transfer knowledge among deep neural networks for the purpose of solving a task of Part-of-Speech tagging for relatively close languages and in the process help each other.

[1]http://labs.theguardian.com/digital-language-divide/

## 1.3 Research Questions

This thesis focuses on the transfer learning capability between Hindi and Nepali language. The primary goal of this thesis work is to:

**To investigate whether a deep learning model, when trained jointly on the Hindi and Nepali language, can help improve its performance on common a task in both the languages.**

Hence, the thesis tries to explore the answers to the following questions:

- *Can embeddings trained in both Hindi and Nepali languages help a deep learning model perform better on common tasks in both the languages?*
- *Can transfer learning between Hindi and Nepali language help a model perform better on the task of POS tagging in both languages?*
- *Can multitask learning in Hindi with the main task of POS tagging and auxiliary tasks of gender tagging and singularity/plurality tagging, help in the improvement of the performance of the model in the task of POS tagging?*
- *Can vector mapping the word embeddings of Hindi and Nepali cross-lingually help in improving the performance of the model?*

## 1.4 Thesis Contribution

In this thesis work following tasks were accomplished:

1. Cross lingual word embedding mapping for Hindi and Nepali word embeddings.
2. Part of Speech tagging for Hindi language using BLSTM-CNN-CRF architecture.
3. Part of Speech tagging for Nepali language using BLSTM-CNN-CRF architecture.
4. Multitask learning in Hindi language for the task of POS tagging, gender classification and singular/plural classification by modifying BLSTM-CNN-CRF architecture.
5. Transfer Learning between Hindi and Nepali language for the task of POS tagging for both the languages by modifyig BLSTM-CNN-CRF architecture.

## 1.5 Structure of this Work

This section gives an overview of the chapters in this thesis work.
The first chapter provides the introduction to the topic and the motivation leading to choosing this topic. It also puts forth some research questions and gives a small overview of the contributions of this work.
Chapter 2 provides an introduction to the basic fundamental concepts to better understand the following chapters. It gives an overview of natural language processing, the task of Part of Speech tagging, and then it goes on to give a brief introduction into deep learning. It also provides a general introduction to transfer learning and deep neural network. Chapter 3 deals with transfer learning and different methods of transfer learning in natural language processing. Chapter 4 provides an introduction to the Hindi and Nepali languages and how are they similar and suitable candidates to explore transfer learning. Chapter 5 provides a brief overview of different implementation in POS tagging in general and then it goes on to explain different POS tagging implementations in Hindi and Nepali.
Chapter 6 gives an overview of the deep learning architecture chosen for the task and explains the function of each components. Chapter 7 provides an overview of the sources of the datasets used in this thesis work and provides the architectural overview into the implementations of individual POS tagging for Hindi and Nepali languages, multitask learning in Hindi and transfer learning between Hindi and Nepali for the task of POS tagging. Chapter 8 provides information regarding the hardware and software used for the implementation and the metrics used for the evaluation of the performance of the model. Then it goes on to explain the results with a discussion.
Finally, chapter 9 provides a conclusion of this work by trying to answer the research questions put forth at the beginning of this work and it also provides possible future work.

# 2 Fundamentals

This chapter provides an overview to the fundamental concepts required to better follow through this thesis work. It first provides an introduction to natural language processing and the task of part-of-speech tagging. Then it moves on to explain a general concept of deep learning and transfer learning.

## 2.1 Natural Language Processing

The phrase natural language processing(NLP) has already been in use since around the 1950s. It basically deals with altering unstructured language data used by human beings in such a way that computers can comprehend it. Wikipedia defines Natural language processing as (see [2]):
**"a subfield of linguistics, computer science, information engineering, and artificial intelligence concerned with the interactions between computers and human (natural) languages, in particular how to program computers to process and analyze large amounts of natural language data."**
Computers can be enabled to comprehend human language by building an interface between languages and computers. Hence, NLP can help us in implementing various applications in the real-world scenario which can provide a possibility of solving complicated tasks.
Even though, it may appear rather theoretically straightforward but building a natural language processing model based on real-world scenarios is actually a challenging task. A few of the reasons are the lack of clarity, disorganizedness, and inaccuracy. It is sometimes difficult even for normal people to comprehend the proposed interpretation of a given sentence. These kinds of sentences are known as syntactically ambiguous. Given below are three examples of such sentences(examples are from [4]):

- *"The professor said on Monday he would give an exam."*
- *"The chicken is ready to eat."*
- *"The burglar threatened the student with the knife."*

Sometimes the intended meaning of the sentences can be derived by sheer reasoning, i.e. in the third example, it is more likely that "the burglar having knife threatened the student" than "the student with the knife was threatened by the burglar". However, it is always difficult to understand the ambiguous sentences as in the above examples and to get the entire essence of what the author intended. The ambiguity of the natural language is also further depicted in figure 2.1 in which it could be seen that the same sentence *"Fly over the boat with the red bow"* can be interpreted in four different ways.

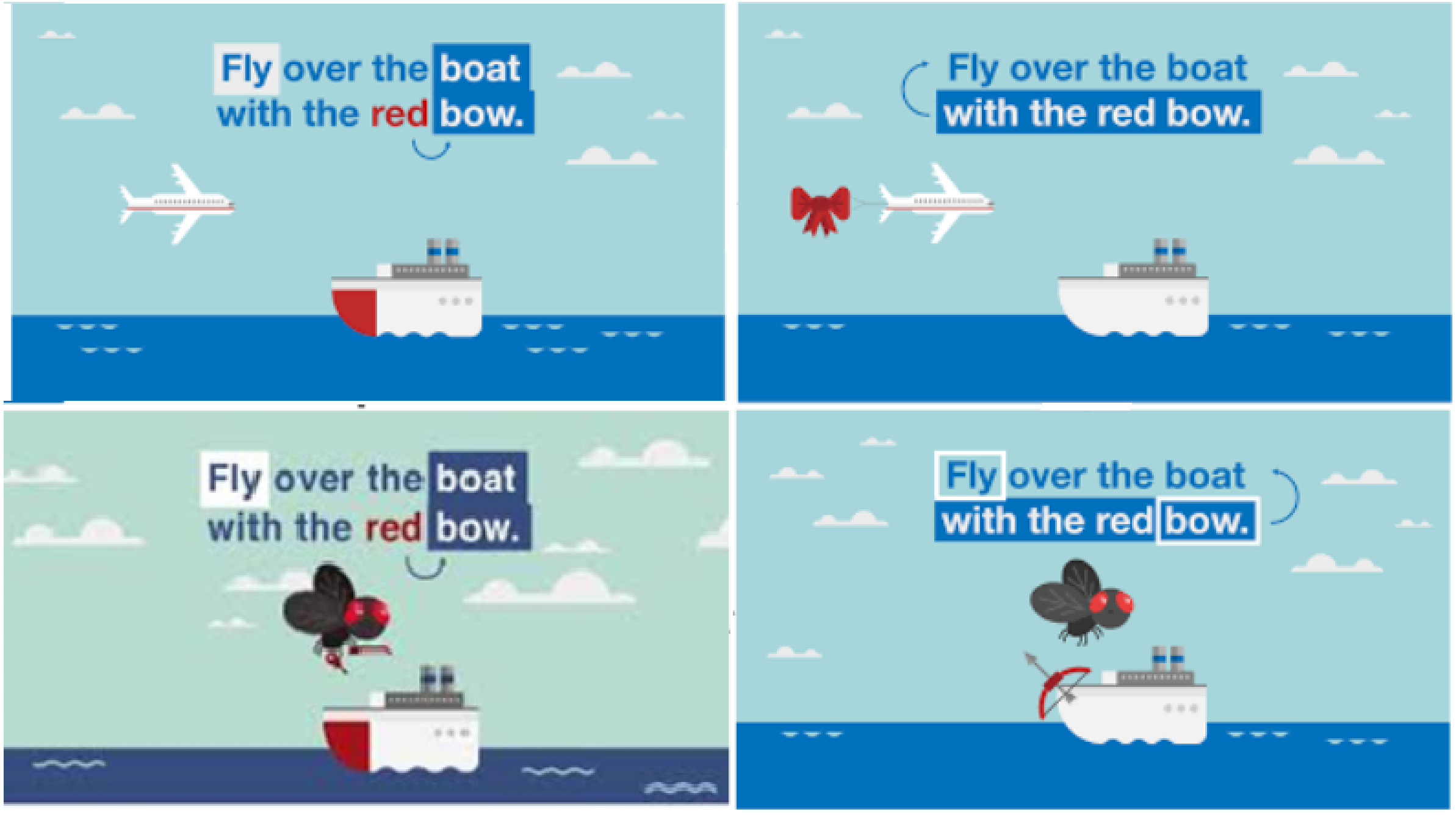

**Figure 2.1:** Ambiguity in natural language processing (from [5]).

It signifies the obvious difficulty in implementing this kind of knowledge base in machines. Nevertheless, rigorous research in this domain has led to the development of many state-of-the-art algorithms that performs reasonably well on various tasks.

This following section derives it's knowledge from the lecture on natural language processing given by Kristina Yordanova and Frank Krueger at the University of Rostock. Figure 2.2 shows tasks in natural language processing which are namely, preprocessing, syntactic analysis, semantic analysis, and contextual interpretation. The preprocessing phase is further subdivided into tokenization and normalization, part-of-speech tagging, morphological analysis, named entity recognition, and co-reference resolution. These terms are explained below briefly.

- **Preprocessing**
  The phase of preprocessing in natural language processing contains following tasks (see [6]):
  - *Tokenization and Normalisation*
    The task of identifying word and sentence boundaries is called tokenization. The letter like period(.) could mean at some place the completion of a sentence or it could also signify abbreviation. Normalization basically means

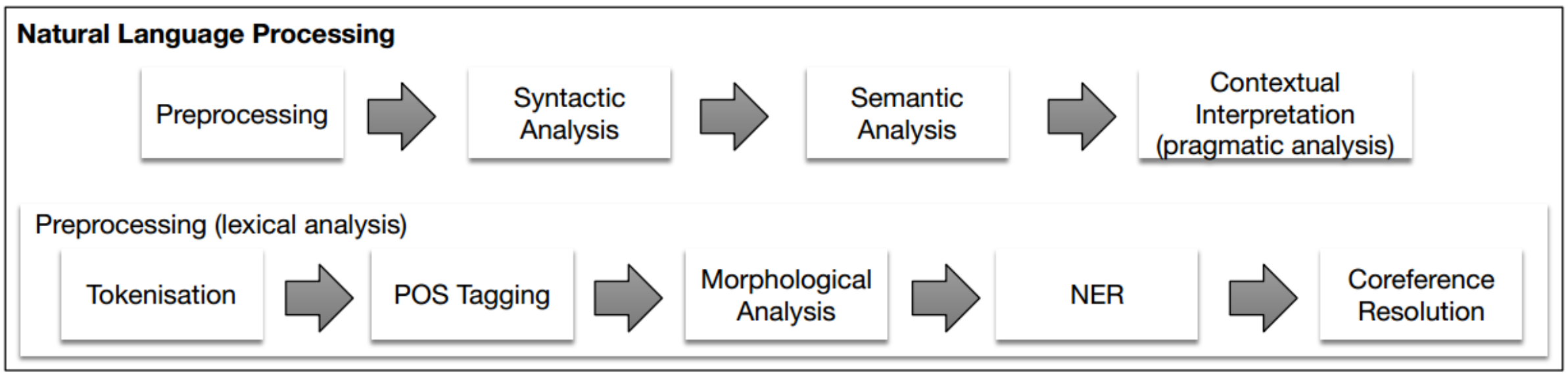

**Figure 2.2:** Tasks in natural language processing (from [6]).

finding different words which could mean same thing and transforming them in homogeneous formats. For example: 10th of July,2020 and 10/07/2020 would mean the same date.

- *Part-Of-Speech(POS) tagging*
  As the name signifies, this task basically assigns to each word it's respective part of speech [6]. It is also further explained in section 2.2.
- *Lemmatisation*
  This operation maps the words to it's root form. For example, words 'Best' and 'Better' would be mapped to 'Good'.
- *Named Entity Recognition(NER)*
  It identifies the words in unstructured text which implies unique objects and classifies them into a pre-defined classes like person, location, organizations,etc. For example, 'Mount Everest', 'Nepal', 'Dell',etc.
- *Coreference Resolution*
  It identifies if two or more different entities means the same exclusive entity. For example, the word 'U.S.A.', 'U.S.' and 'the U.S.' refers to the same country.

• **Syntactic Analysis**
Syntactic analysis is mostly focused on exploring the syntactical structure of the given sentence. It is further categorized into parsing and chunking. Parsing deals with exploring the syntactical structure of the sentence whereas chunking deals with grouping together words that fall under a similar category of syntactic meaning.

• **Semantic Analysis**
Semantic analysis is concerned with linguistic sentence meanings. It is further subdivided into semantic interpretation and ambiguity resolution. Semantic interpretation deals with the word meanings whereas ambiguity resolution deals with exploring precise word meanings.

• **Contextual Interpretation**

Contextual Interpretation is concerned with exploring the essence of a sentence based on the knowledge of the context and the knowledge of the word. For example, the pronoun ”He” in ”He is traveling to Berlin” implies that it refers to a boy but this knowledge can’t be extracted from the given sentence.

The real world applications of NLP include automatic text summarization, sentiment analysis, machine translation, information retrieval, etc.

## 2.2 Part-of-Speech tagging

Part of speech tagging deals with labeling each word in a sentence with the probable part of speech it would belong. It is considered as a syntactic analysis of the language at the most basic level. An example of it is shown in figure 2.3 which uses the Stanford university part-of-speech tagger [7]. It is helpful in disambiguating the word sense and also parsing syntactically. [6]

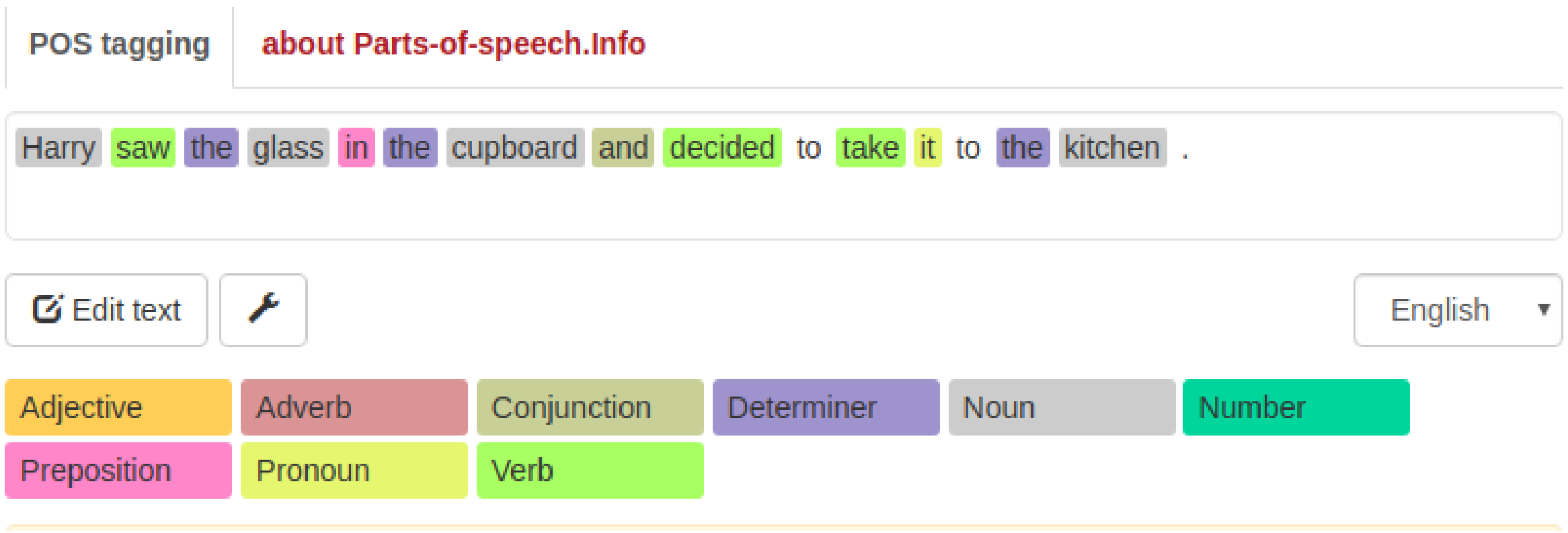

**Figure 2.3:** Part of speech tagging online by (see [7]).

The major POS tagsets available in English are:

- The British National Corpus[1] containing 61 tags.
- Pen Treebank Corpus with 45 tags (see [8]).
- Brown Corpus[2] with 87 tags.

### 2.2.1 Parts of Speech tags in English

The major English part of speech tags according to Penn Treebank are (see [8]):

[1]https://www.english-corpora.org/bnc/
[2]http://korpus.uib.no/icame/manuals/BROWN/INDEX.HTM

- Noun
  It is used to specify a particular person, thing or place. The various tags in Noun are:
  - Singular(NN): cat, spoon.
  - Plural(NNS): cats, spoons.
  - Proper (NNP, NNPS): Luke, Rostock.
  - Personal pronoun (PRP): I, me, she.
  - Wh-pronoun (WP): who, whom, what.
- Verb
  It signifies any action or processes. The various tags in Verb are:
  - Base, infinitive (VB): sit.
  - Past tense (VBD): sat.
  - Gerund (VBG): sitting.
  - Past participle (VBN): sat.
  - 3rd person singular present tense (VBP): sits.
  - Modal (MD): could, would.
  - To (TO): to (to sit).
- Adjective
  An adjective is a word and Part-of-Speech which modifies the Noun. The various tags in Adjective are:
  - Basic (JJ): green, short.
  - Comparative (JJR): greener, shorter.
  - Superlative (JJS): greenest, shortest.
- Adverb
  The adverbs modifies verb. The various tags in Adverb are:
  - Basic (RB): fastly.
  - Comparative (RBR): faster.
  - Superlative (RBS): fastest.
- Preposition (IN): under, on, with.
- Determiner:
  - Basic (DT): a, an, the.
  - WH-deteminer(WDT): whose, which.

- Coordinating Conjuction (CC): and, or, that.
- Particle (RP): on (put on), up (put up).

### 2.2.2 Techniques of POS-tagging

There are basically two approaches of part of speech tagging (knowledge from [6]).

- **Rule-Based:** This method replies heavily on rules crafted by humans on the basis of lexical and other language knowledge.
- **Learning-Based:** This approach basically deals with various models as given below which learns POS-tagging from manually labeled corpus such as Penn Treebank.
  - Statistical models: Hidden Markov Model (HMM), Conditional Random Field (CRF).
  - Rule learning: Transformation Based Learning(TBL).
  - Neural networks: Long Short Term Memory(LSTMs).

## 2.3 Deep Learning

This section gives an overview of essential deep learning concepts that are needed to get a clear understanding of the implementation of this thesis work. It will first provide a primary explanation on the key topic of deep learning and neural networks and then it will proceed with a more detailed explanation of transfer learning as the basic understanding of this concept is essential for this thesis work. For further in-depth explanation please refer to: [9].

### 2.3.1 An Introduction to Deep Learning

From the time since computers have been in general use, various methods have been explored to make them smarter and eventually empower them to become artificially intelligent. Initially, it began with the trial to solve those problems that were hard to solve by humans but can be formulated effectively using mathematics. However, over time, the tasks that could be easily solved by humans just by their intuition became the main focus of these intelligent machines. The problem with these tasks is that it is not easy to describe them formally. Some examples of these kinds of tasks would be the detection and recognition of objection in images/videos or speech recognition. (knowledge from [9])

Deep learning is a type of machine learning which is a part of the umbrella concept of artificial intelligence. Deep learning facilitates the construction of intricate concepts using plain components. We will dive into a little bit more detailed explanation of deep learning with the help of figure 2.4. This example with object detection in image was chosen for the purpose of simplicity in explanation.

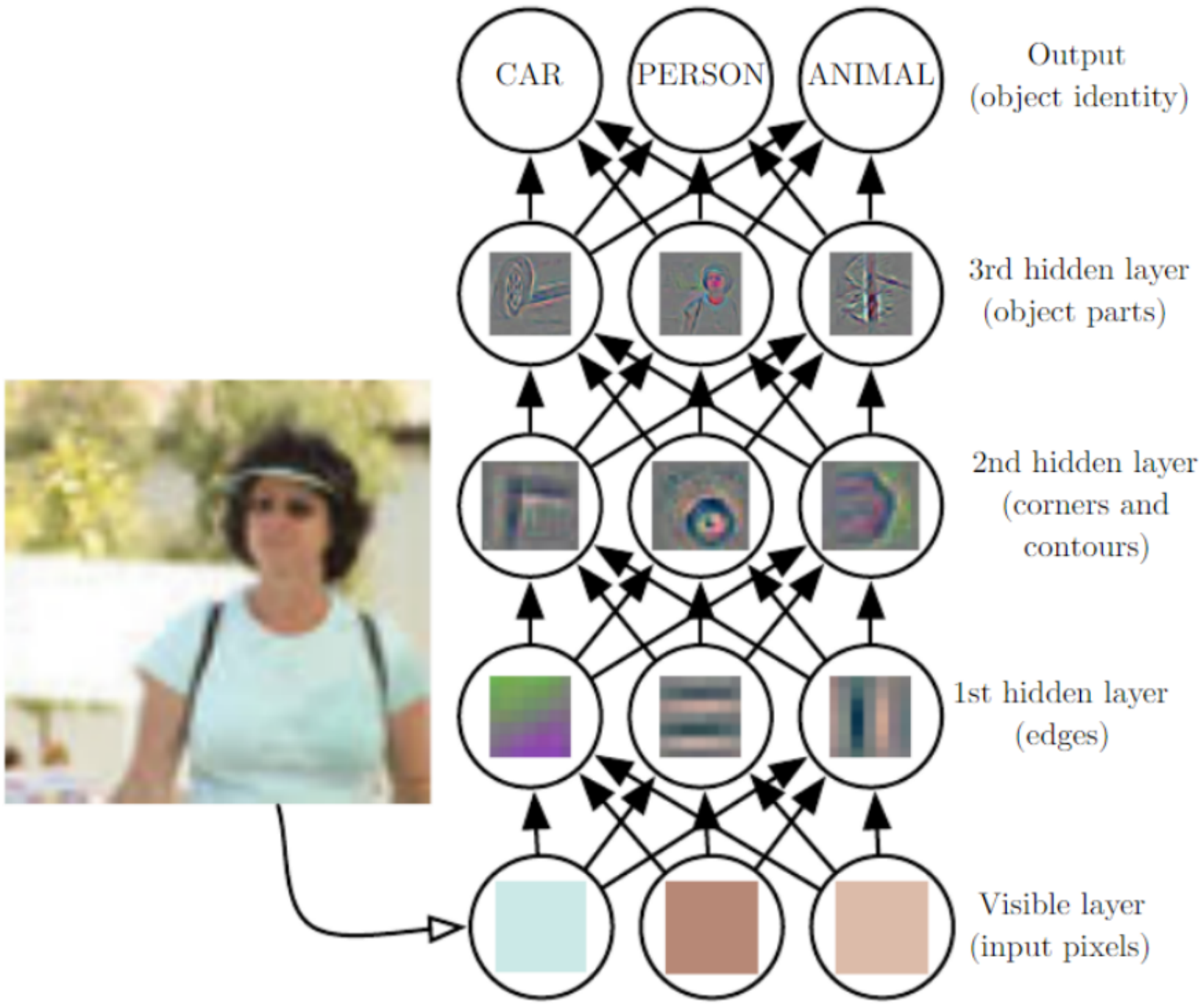

**Figure 2.4:** Image recognition using deep learning (from [9]).

An image is fed as input data to the algorithm which is known as the neural network. The purpose is to detect and identify the pixels in the image. This can be achieved only by learning mappings between pixels and the correct object in the image. It is very difficult to achieve this kind of mappings, in some cases, it is kind of impossible to get it, and hence there is a need for a different way. The problem is addressed in neural networks by breaking it down into multiple layers with simple feature mappings.

As it can be seen in figure 2.4, initially the image at the input is provided with pixel representation. These raw images which are fed to the neural network are called visible or input layers. On the basis of the quality of input image, the hidden layers extract relevant features from the image as it passes through more number of hidden layers, and as it passes through the hidden layers, it also leads to the increase in the complexity of the representation. The reason why these layers are called hidden is that the value that they possess is solely decided by the algorithms and is not known outside. As the image passes through each layer, the layer outputs the extracted image features.

An exact pipeline exhibiting the working of a neural network for image recognition can be as follows. The neural network at the input layer is fed with images as input data. In the second step, the basic features like brightness, edges are detected and extracted by the first hidden layer. In the second hidden layer, the task of identifying the corners and contours are done by clustering together the output edges by the first hidden layer. Now, the next hidden layer takes the corners and contours as the input and attempts to deduce a characterization of the objects or parts of the objects in the image. And finally, in the output layer, the algorithm uses the identified parts of the object from the last hidden layer to conclude a portrayal of the image and responses with the identification of the object.

## 2.3.2 Neural Network

A neural network is a basically combination of multiple artificial neurons connected together. Figure 2.5 shows a single neuron in a neural network. Here, $x_1$, $x_2$, $x_3$ are inputs and each input is multiplied with specific weights $w_1$,$w_2$, $w_3$ respectively. As it can be seen, we have a mathematical operation of summation, $\sum$ and a mathematical activation function, $\sigma$. All the inputs are multiplied with their respective weights and then added together at the summation block. And an extra value called bias ¸is added to this summation which helps in tuning the output And the output of this summation block is fed to the non-linear activation function. Activation function in a a neuron defines the output given the input. There are various kinds of activation functions like tanh, ReLu, Sigmoid, etc. For further information, please refer to [9].

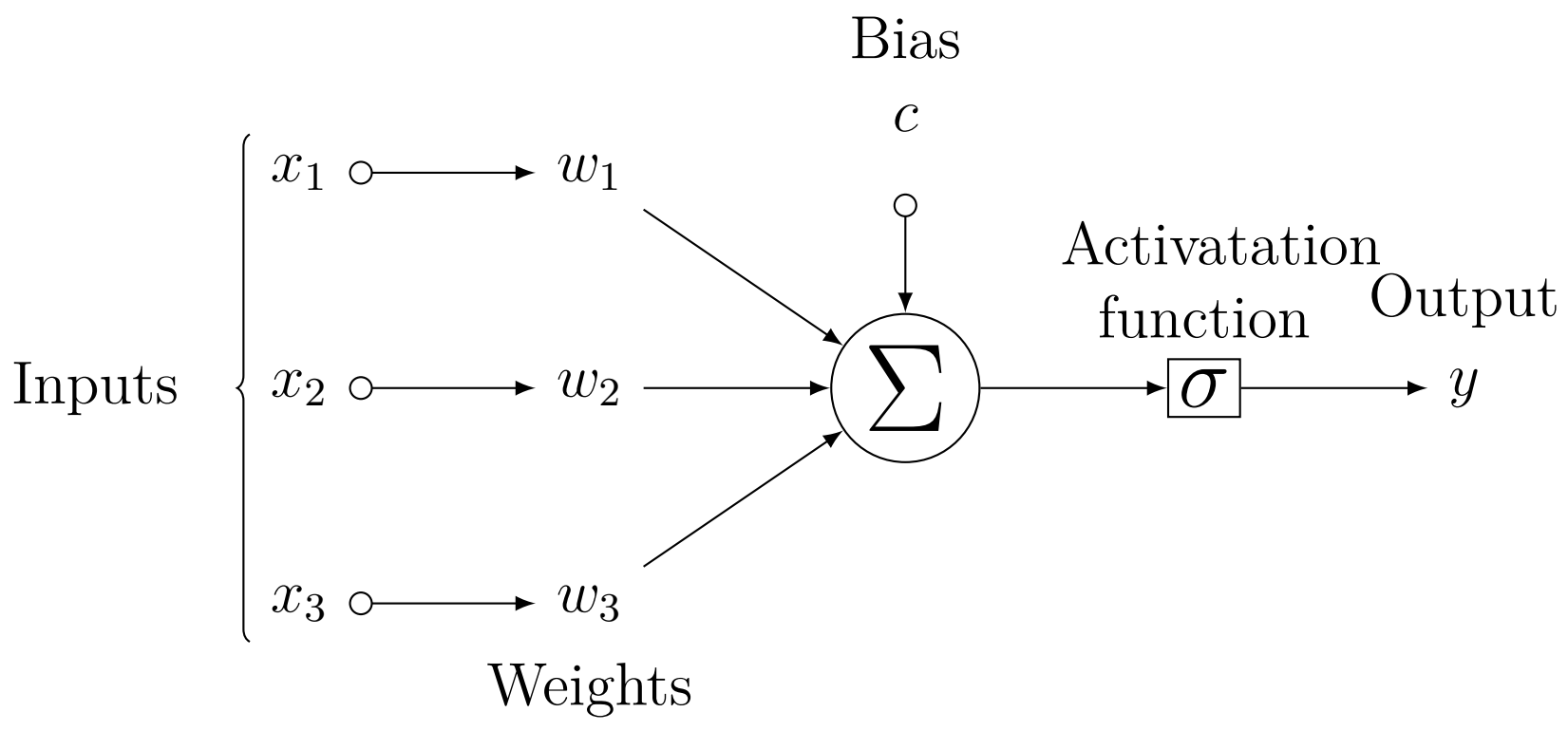

**Figure 2.5:** A neuron in the Neural Network [10].

As, it can be seen in figure 2.6, it shows three layer neural network with first layer as input layer then the hidden layer and finally the output layer.Each node in this neural network signifies an artificial neuron. Input is fed to the initial layer and then it passes on to the hidden layer and finally to the output layer. At each layer, the computations on the input are performed and further passed on to the next layer and then finally the output at the final layer.

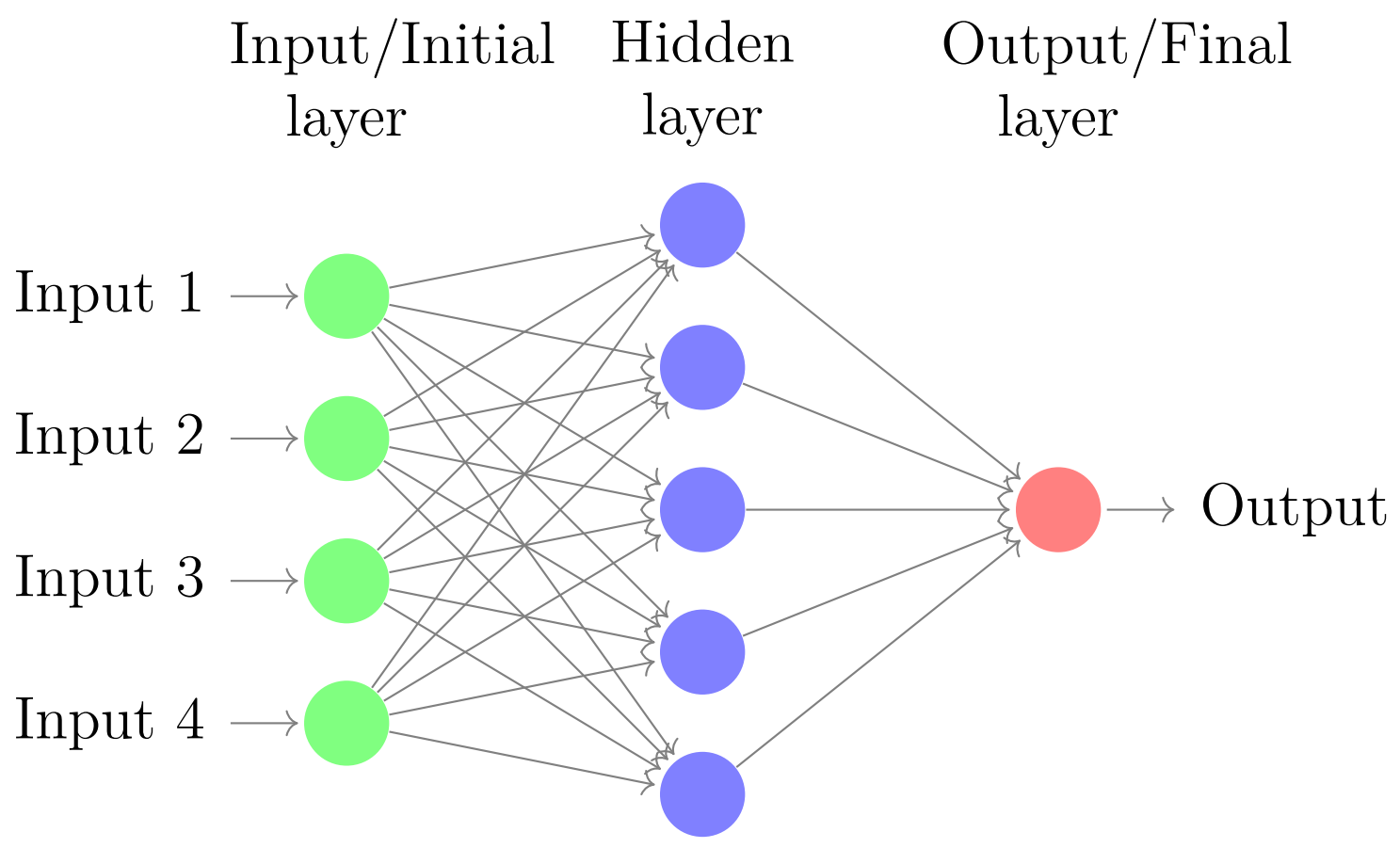

**Figure 2.6:** An example of neural network [11].

### 2.3.3 Transfer Learning

According to the *Deep Learning* book by Ian Goodfellow and Yoshua Bengio, Transfer learning is defined as:
"**Transfer learning and domain adaptation refer to the situation where what has been learned in one setting . . . is exploited to improve generalization in another setting**"(see: [9])
This subsection gives an overall introduction and features various methods of transfer learning. It first gives a brief definition, then moves on to potential scenarios and the possible benefits of transfer learning.

#### 2.3.3.1 Introduction to Transfer Learning

Deep learning covers a vast spectrum of topics in the field of artificial intelligence. The topic of central importance in this thesis work is transfer learning.

In the conventional method of supervised machine learning, the builder of the neural network wants the network to perform well on a particular task. And there is a need for a huge amount of annotated data in order to achieve better performance. These annotated data belongs to a particular place which is called domain. To achieve better accuracy, there is a need for enough data which can help the model to adjust its weights appropriately. In a lack of enough data, the model would most likely not perform well. Now, this kind of problem is more prevalent if the weights and biases of the model are randomly initialized for the training of the model. In figure 2.7, the left-hand side shows the probable way in which a network can come to a final solution when it doesn't possess any previous familiarity with the problem. This particular way of learning in which the network discovers the rules by looking at the examples is called inductive learning. It is quite evident that the network would need a lot of backpropagations to update its weights to finally arrive at the optimal solution. And for the purpose of updating weights at every backpropagation, a huge amount of annotated dataset is necessary if it is to be in the right direction. Backpropagation basically is an algorithm used to adjust the weights and biases of the neural network. As the network trains on more amount of annotated datasets in a domain, it would eventually perform well for this particular task in the respective domain. Hence, this approach could lead to good accuracy for tasks in the same domain given enough labeled data. (knowledge source: [12])

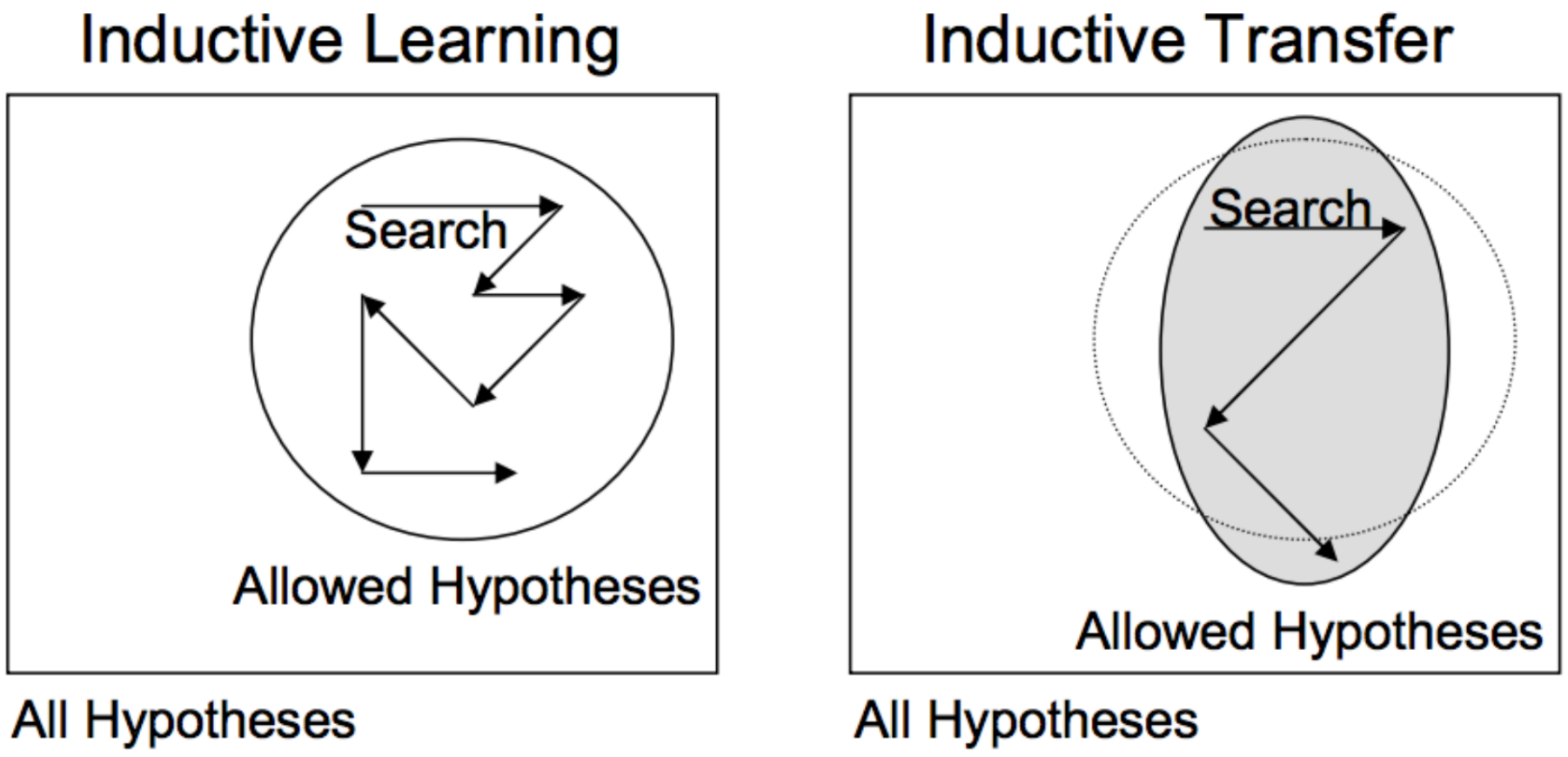

**Figure 2.7:** Inductive learning and transfer learning (from [13]).

However, the advantages of the above approach are limited only to a task in one domain, i.e., if the same trained neural network is used to perform tasks that are even slightly different than the original task, it would probably perform worse. For example, if we train and use an object detection and recognition network for the detection of objects like dogs, cats, etc. in broad daylight, and again if we use the same network to detect these objects in say, a foggy or cloudy day with bad lighting conditions, the chances are really high that it would perform worse just by little change in the lighting conditions. The main reason for this difference in performance is that during the training of the neural network, all the weights and biases which affects its output were adjusted only on the images from the daylight. As the network's weight were not optimized on the images other than daylight, it was not able to generalize well and couldn't perform well on the images with slight difference in domain. (knowledge source: [14])

Transfer learning can help us in solving the problem mentioned in the previous paragraph without having to train a completely new neural network with a lot of annotated data from the new domain, in this case pictures from the foggy or cloudy day with poor visibility. In transfer learning, a neural network is trained on in some domain say A to solve a task in the same domain A and then the same neural network is used to solve a task in other domain say B by training on a small amount of annotated data in domain

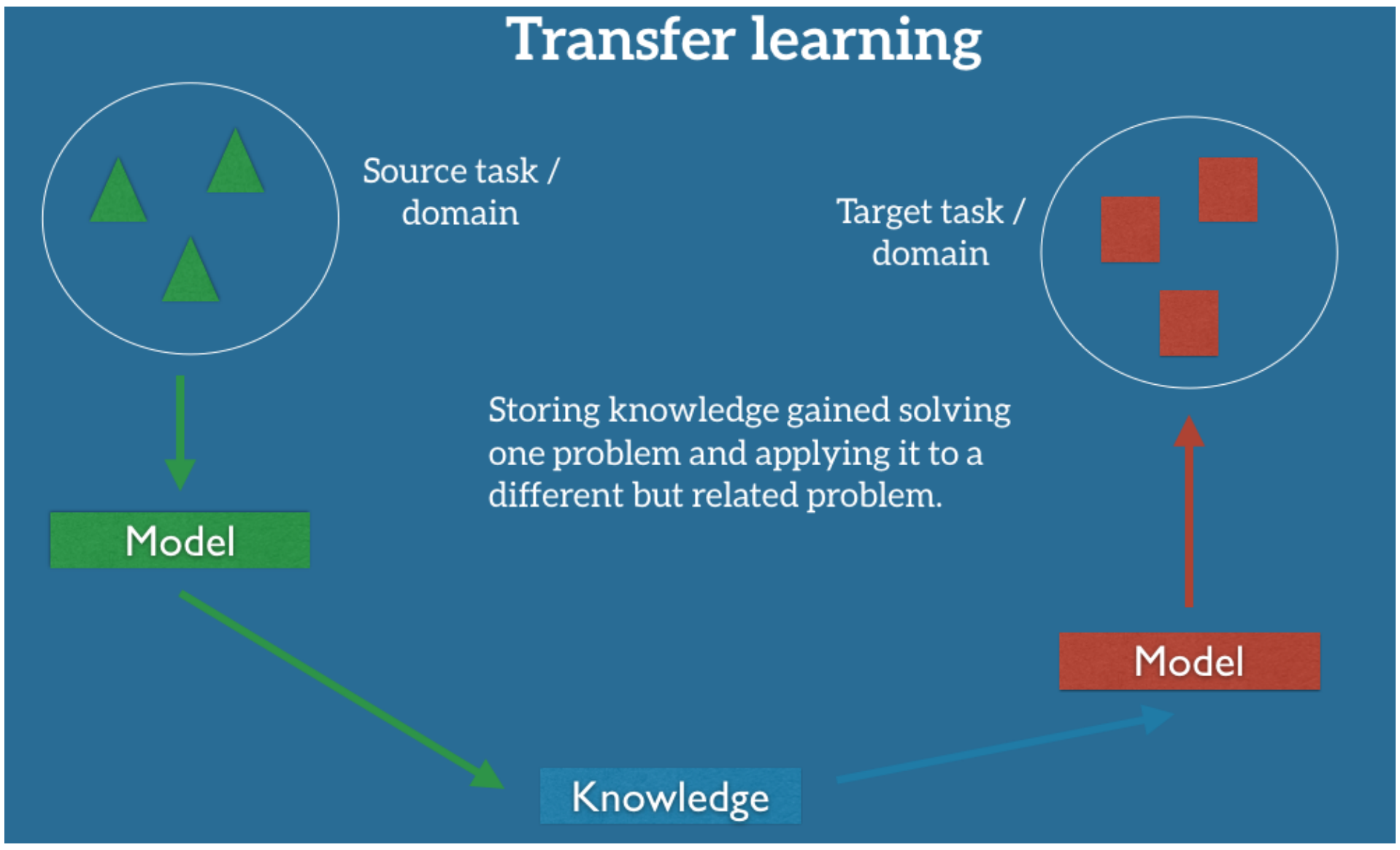

**Figure 2.8:** A general setup in Transfer Learning (from [14]).

B. This is also called inductive transfer as shown in the right-hand side of figure 2.7. The fundamental setup of transfer learning is depicted in figure 2.8 where it can be seen that a model is first trained on data from one source task/domain and then the accumulated knowledge is used on target task of different task/domain (for more information refer to: [14]). However, the main goal of transfer learning is not just to train a model in one domain and use it in a new domain but also to help the model generalize better on the features from both domains so that it can perform better on the tasks from both the domains.

#### 2.3.3.2 Possible Modes to Consider Transfer Learning

This subsection deals with possible modes in which transfer learning could be considered and these categories are inspired by [13]. All the four possible modes are shown in figure 2.9. It basically depicts a plot of the size and data similarity of the new task given a neural network is previously trained on some other task. All the possible scenarios are discussed below:

1. *Abundance of data & high resemblance:*
   It is the best possible scenario in transfer learning where we have a huge amount of labeled data and the data is kind of similar to the original labeled data on which the model was pre-trained. In this state, we don't need to let go of any of the knowledge gained by the algorithm from the previous task as the features are kind of similar and neither is there any need to change the weights of layers of the pre-trained model completely. The only thing needed to be done in this mode is to fine-tune the pre-trained model on the new dataset to make it learn the slightly new features by adjusting the weights and biases.

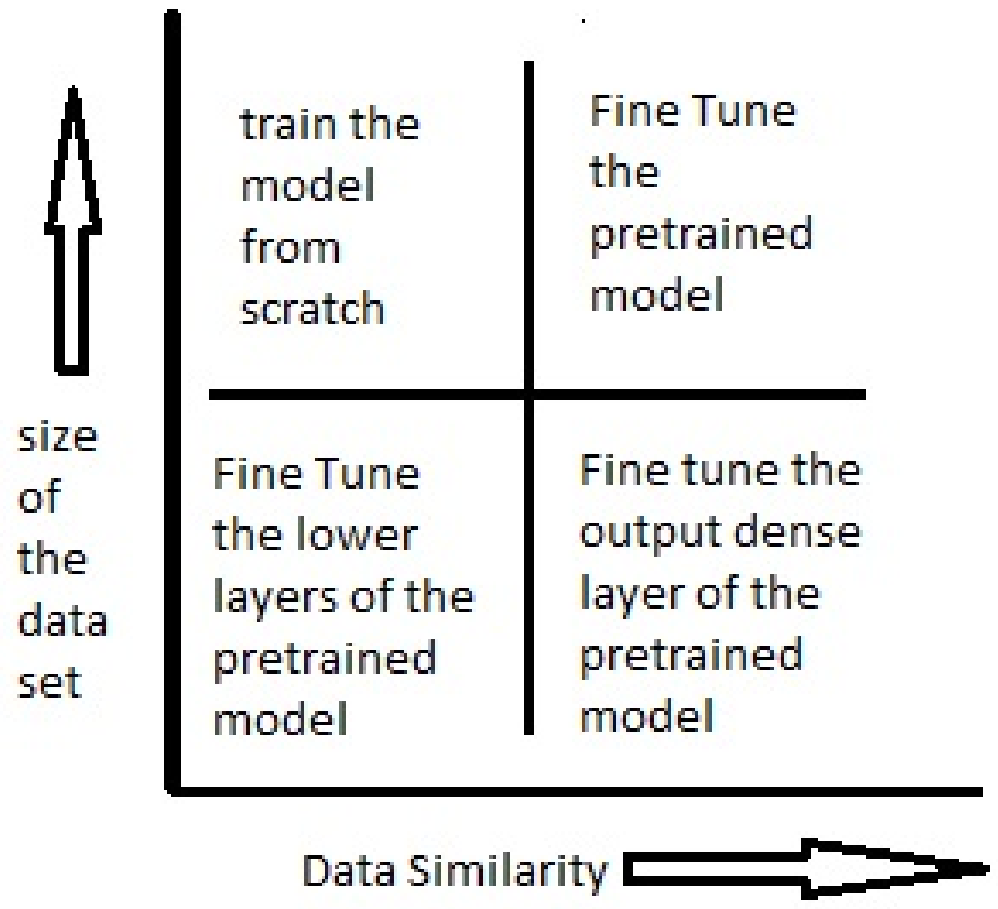

**Figure 2.9:** Modes in Transfer Learning(from [13]).

2. *Abundance of data & low resemblance:*
   In this case, the amount of labeled data is high however the similarity of the data between the new dataset and previously trained dataset is low. The best possible approach in this condition would be to train a new neural network from scratch. The reasons for that are:
   - The new network doesn't need the transfer of knowledge from the original dataset as it has enough dataset to learn from scratch.
   - And, when the pre-trained network is trained on the new dataset, it could probably perform worse as the previous weights and biases are trained in different directions.

3. *Scarcity of data & low resemblance:*
   This case is least considerable in terms of transfer learning. The only way to use transfer learning, in this case, is to save the weights of the lower layers in the neural network as they store the basic features and to fine-tune the later layers specific to the task. The main reason for this is, to keep lower layers that are able to retain the ability of feature extraction similar in both datasets and then train the higher layers specific to the task. But as it can be guessed, it doesn't provide good results.

4. *Scarcity of data & high resemblance:*
   In this case, the new dataset is not enough for the neural network to train from scratch and provide better performance. However as the new and the old dataset are kind of similar, it is highly plausible that just by fine-tuning little bit the output layer, there are great chances that the network can provide good performance on the new dataset. (see [13])

#### 2.3.3.3 Advantages of Transfer Learning

According to a survey done by Pang et. al. [15] on transfer learning, there are various implicit benefits of using a neural network to accomplish more than one similar tasks. The three main advantages are:

- *Superior initiation:* As the neural networks learn its weights and biases from the previous task, it reaches better performance on the new task even after few epochs. Hence, it has a head start over the models which start training on the new dataset from scratch.
- *Better learning curve:* The neural network shows a better learning curve and reaches higher performance faster compared to the networks learning from scratch.
- *Better performance:* And the neural network pre-trained on a similar task seems to perform better reaching higher accuracy than the models training from scratch as the weights are more generally biased.

Hence, if the datasets are abundant and if there is a similarity between tasks to be performed by neural networks then there is a higher probabilities that all the above advantages could be reaped. Figure 2.10 shows the possible advantages of transfer learning. As mentioned in the advantages, the neural network with transfer learning shows better performance with higher slope, higher start, and higher asymptote (for more information, refer to [16]).

## 2.4 Word Representations

The words being fed to the deep neural network has to be numerically represented so that the various mathematical operations in the neural network can be performed. However, as we know that words in natural language have relationships between them, and some words have more similarity between each other than others. For example, words like 'Berlin' and 'Rostock' both represent cities in Germany, words like 'tall' and 'long' are more similar to each other. Hence, there is a need to embed this important information that every word contains and word embeddings are a method of storing these important pieces of information in vector representations.
Word embeddings are used for the purpose of word representation where words having similar meanings are represented with similar vector representations. The main advantage of using word embeddings is that it stores the words in low dimensional vectors and it also preserves the syntactic and semantic features of the words. Word embeddings were introduced by Collobert and Weston but the importance of word embeddings in NLP was highlighted by the work of Mikolov et. al. [17] where they used skip-gram models and the Continuous-Bag-Of-Words(CBOW) for the distributed vector representations. The various kinds of word embeddings methods available are Word2Vec, GloVe, fastText, etc.
Word2Vec uses the information of the words in the surrounding of the target word to create a vector representation of the target word using a neural network. There are two kinds of Word2Vec embeddings generation methods, they are skip-gram and CBOW. In skip-gram, the input fed to the neural network is the target word and the output is the words in the surrounding of the target words. The neural network contains just one hidden layer and the dimension of the hidden layer is of the same size as that of embeddings.
CBOW is kind of similar to the skip-gram, however, it interchanges the input and the output. The underlying idea is that if a neural network is given a context, i.e say a sentence, which is the most probable word likely to appear.
Sometimes generating word embeddings could be computationally expensive and hence instead of creating embedding at the word level, the embedding is created at the character level. Character level embeddings have been seen to provide better performance when it comes to deep learning applications on rare languages and it also helps to overcome the problem of out of vocabulary words for unknown words.

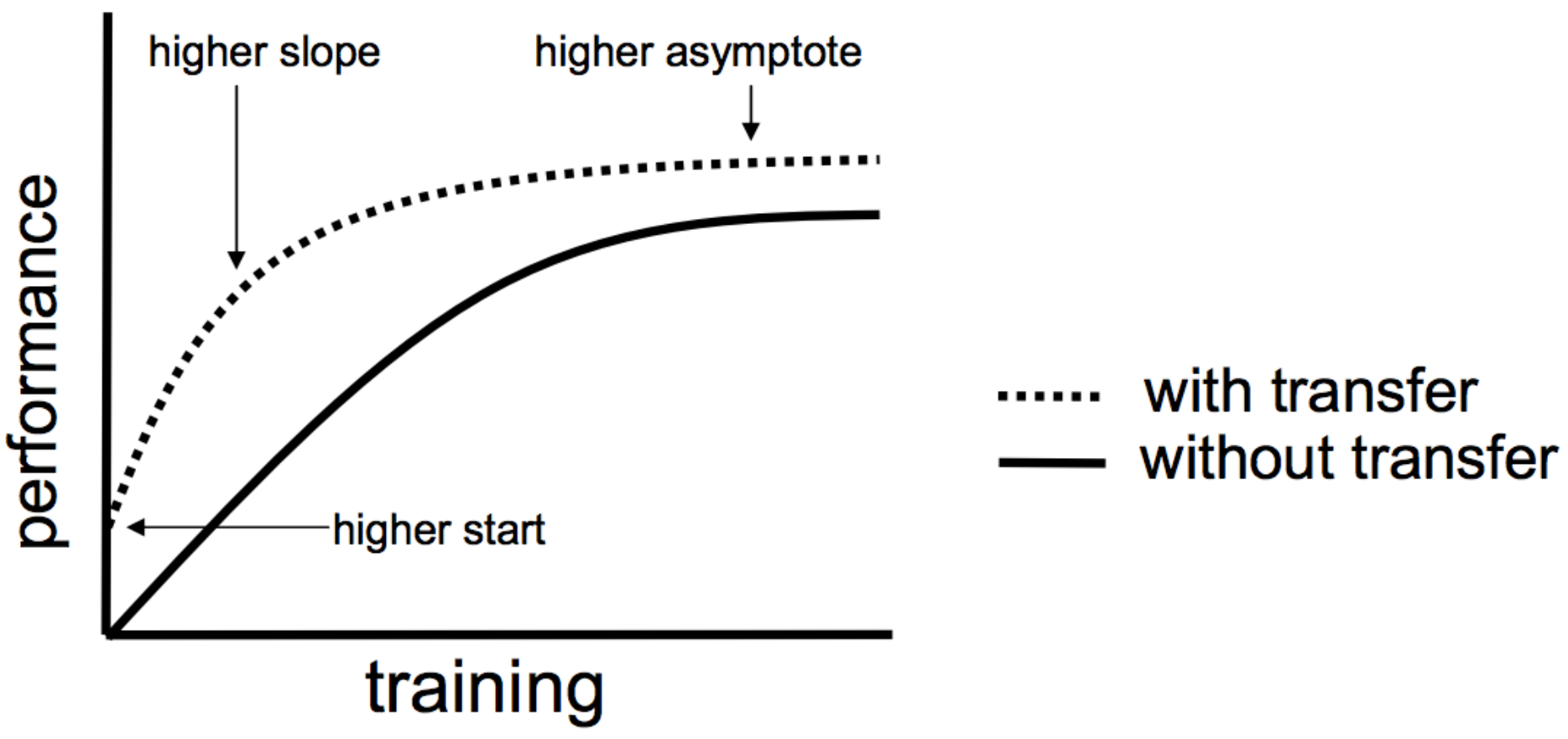

**Figure 2.10:** Advantages of Transfer Learning(from [16]).

### 2.4.1 FastText

FastText was developed by Facebook in 2016 and it is a kind of further development of Word2Vec embeddings. On contrary to feeding individual words in the neural network, it feeds the word after breaking it into various sub-words called n-grams. For example, the tri-grams for the word *Rostock* would be *Ros*, *ost*, *sto*, *toc* and *ock* and ignores the boundaries of words at start and end. So, the word embedding vector for the word *Rostock* will be the sum of all these n-grams in the previous example. Hence after the training of the neural network, the word embeddings of all n-grams are generated. And another advantage of fastText embeddings is that it also takes care of rare words and there is a high probability that some of the n-grams of a rare word can also be found in the n-grams of other words. (please also see: [18] [19] [20])

# 3 Transfer Learning in Natural language Processing

Transfer learning is a part of machine learning which is concerned with accumulating knowledge obtained during working out on one task and using the accumulated knowledge to solve a different but similar task (see [21]). This chapter gives an analysis of research in transfer learning for natural language processing. It examines the possibility of transferring machine learning models to data beyond its present training distribution and specifically across various domains, tasks, and languages. This chapter derives its knowledge from the Ph.D. thesis work of Sebastian Ruder (see: [22]).

## 3.1 Traditional machine learning and transfer learning

In a traditional supervised machine learning scheme as shown in figure 3.1, we would need annotated data from the same domain/task to train a model on some task/domain A. Once, we have trained the model on this particular task then we can suppose that the model would hopefully exhibit better performance on unseen data from the same task and domain. However, when we are provided with new task, say B, from different domain then we need to again train a new model on the labeled dataset to achieve good performance on that task. We would also need sufficient amount of annotated dataset for each task in each domain for the model to achieve high performance and in lack of satisfactory amount of data, the classical supervised models would not perform well.

The above mentioned problem can be dealt with the help of transfer learning. As depicted in figure 3.2, transfer learning is a process where the model A trains for a particular task in domain A and accumulates some knowledge while training for that domain. And then the accumulated knowledge from domain A can be transferred to model B in domain B to solve some different but related task.
3.2 [23].

The knowledge mentioned above could be used to signify various things revolving around the different tasks and domains at hand. However, in this particular work it mostly implies the representation learned by the neural networks.

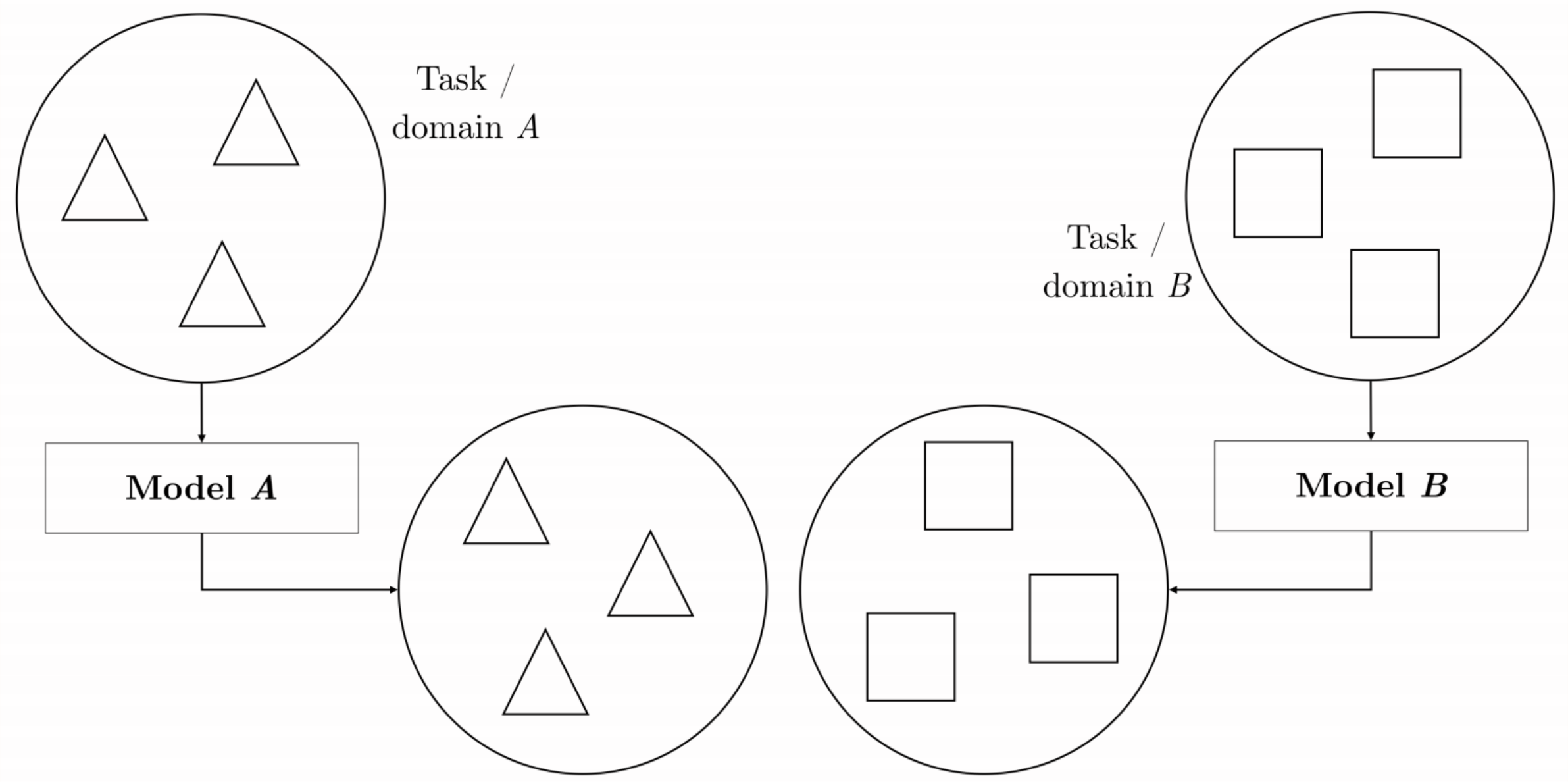

**Figure 3.1:** A depiction of classical machine learning structure [23].

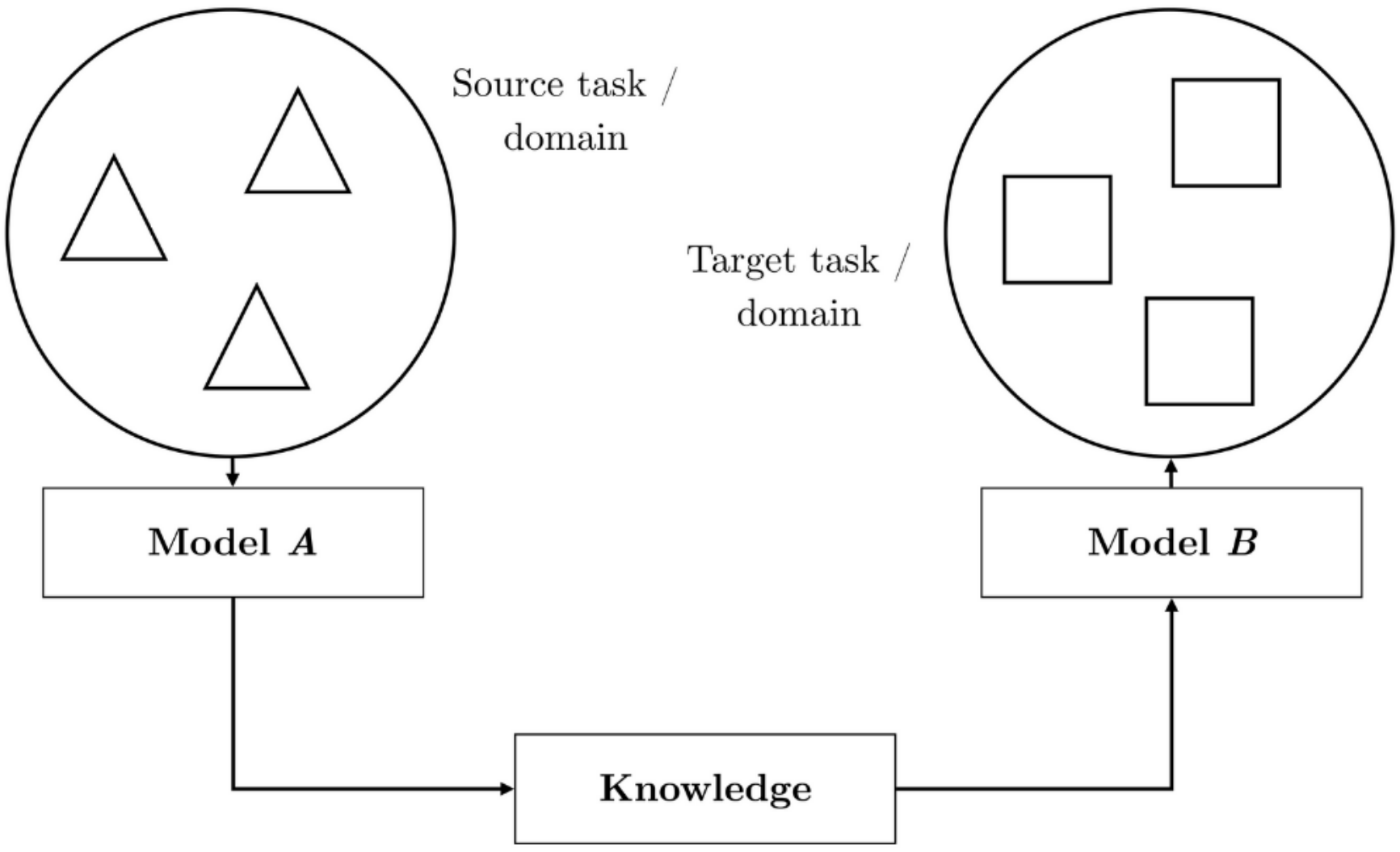

**Figure 3.2:** A depiction of transfer learning [23].

## 3.2 Classification of Transfer learning

This section basically deals with classification of transfer learning for NLP. The whole classification of transfer learning is depicted in figure 3.3. The methodology used here for the classification is on the basis of a) if the source and target model attend to the ditto tasks; b) the sequence of learning the task; and c) the type of source and target domain. If the problems to be solved in both domains are the same but the annotated data is available only in source domain then it is called transductive transfer learning. The transductive transfer learning is further sub-divided into domain adaptation if the tasks occur in different domains and into cross-lingual learning if the tasks occur in different languages. Again, if the problems to be solved are different but if the annotated data is available in target domain then it is called inductive transfer learning which is further sub-divided into sequential transfer learning if the tasks are learned one after another and into multitask learning if they are learned simultaneously.

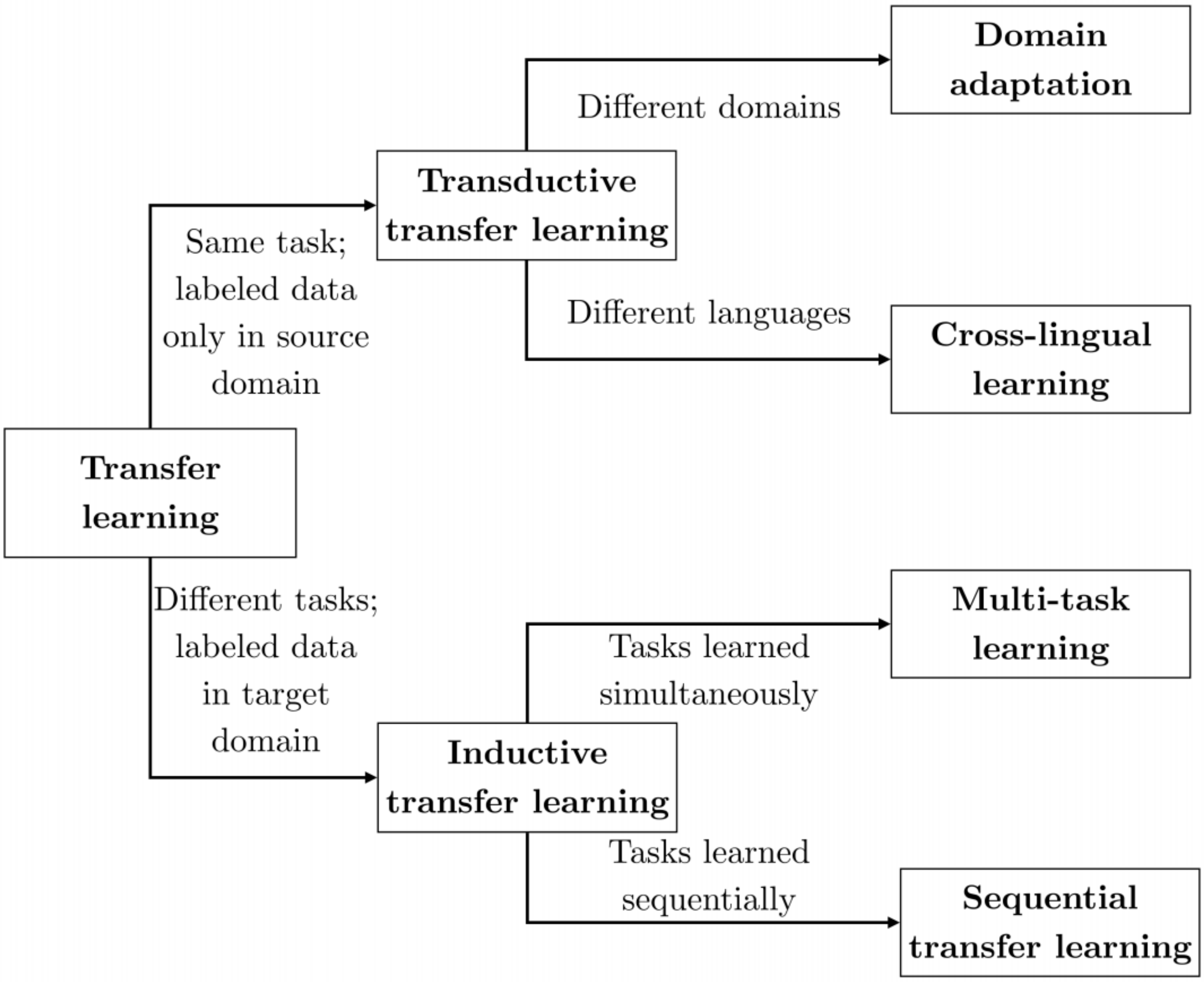

**Figure 3.3:** The classification of transfer learning [23].

### 3.2.1 Multitask Learning

Multitask learning(MTL) is the process of sharing representation between related tasks in order to help a deep learning model in enhancing its generalization on the main task. To be specific, according to Caruna:

**"MTL improves generalization by leveraging the domain-specific information contained in the training signals of related tasks"** [24].
Multitask learning is also called by various names like learning to learn, learning with auxiliary tasks and joint learning.

#### 3.2.1.1 Types of MTL

MTL is typically divided in two types for deep neural networks:

1. Hard Parameter Sharing:
   Hard parameter sharing (as shown in figure 3.4) is the exceedingly used type of MTL. It is implemented by sharing the initial layers of deep neural network for multiple task and then having later task specialized layers separate for each task. Hard parameter sharing has been seen to be helpful with the problem of overfitting as the chances of overfitting is reduced greatly due to the need for the model to learn representations for more than one task. Overfitting basically is a problem when a model learns too well on a given dataset, so much so that it also learns the noise in the dataset and hence has negative impact on the performance of the model. In other words, it memorizes the data instead of learning the rules.

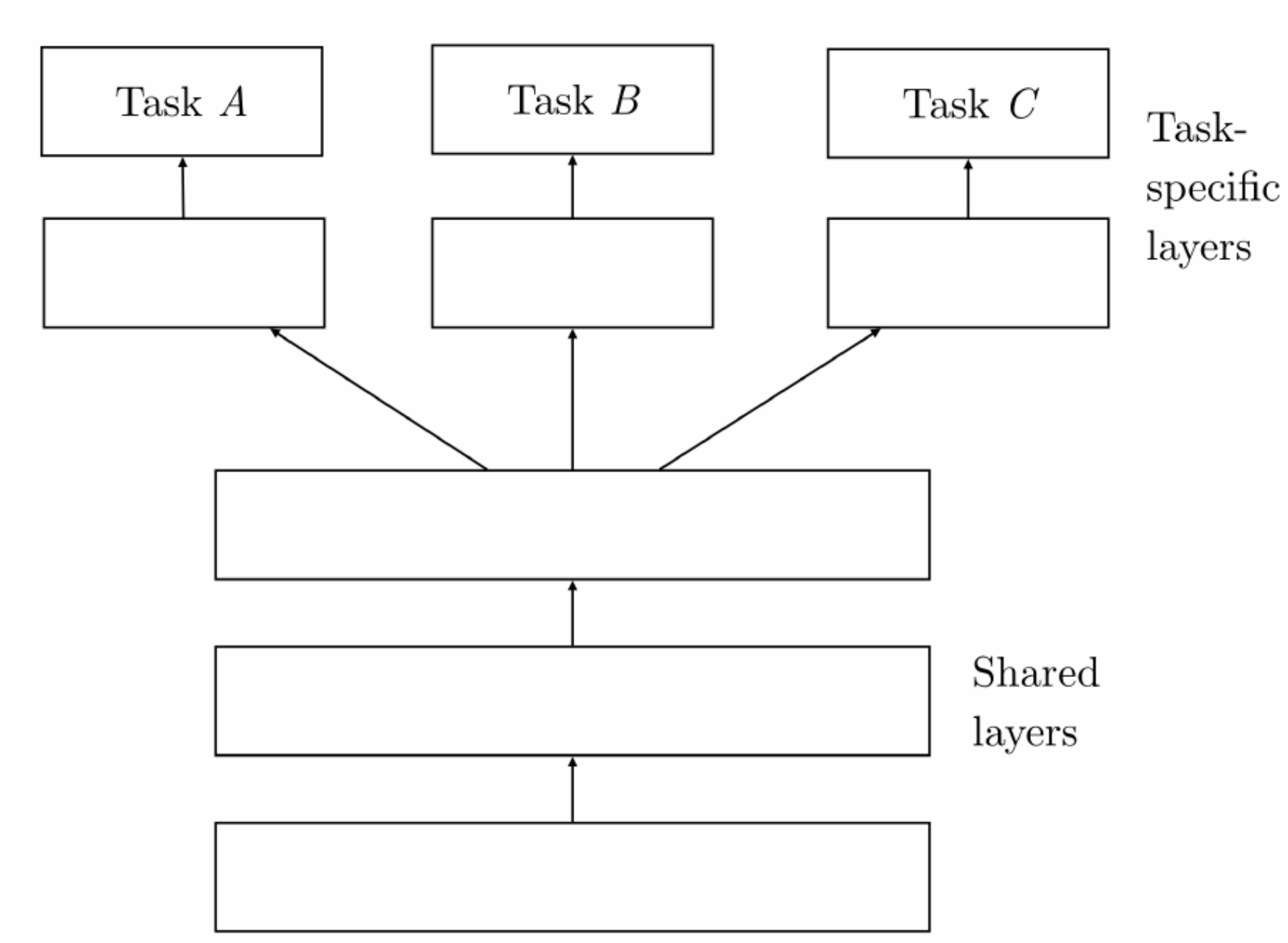

**Figure 3.4:** Hard parameter sharing in multitask learning [23]

2. Soft Parameter Sharing:
   Soft parameter sharing (as illustrated in figure 3.5) is concerned with regularizing the gap between parameters of individual tasks with individual parameters with the goal of keeping them analogous. Regularization is a technique used in deep learning model in which there are a small modifications made to the learning algorithm in order to let the model generalize better on the dataset and avoid overfitting.

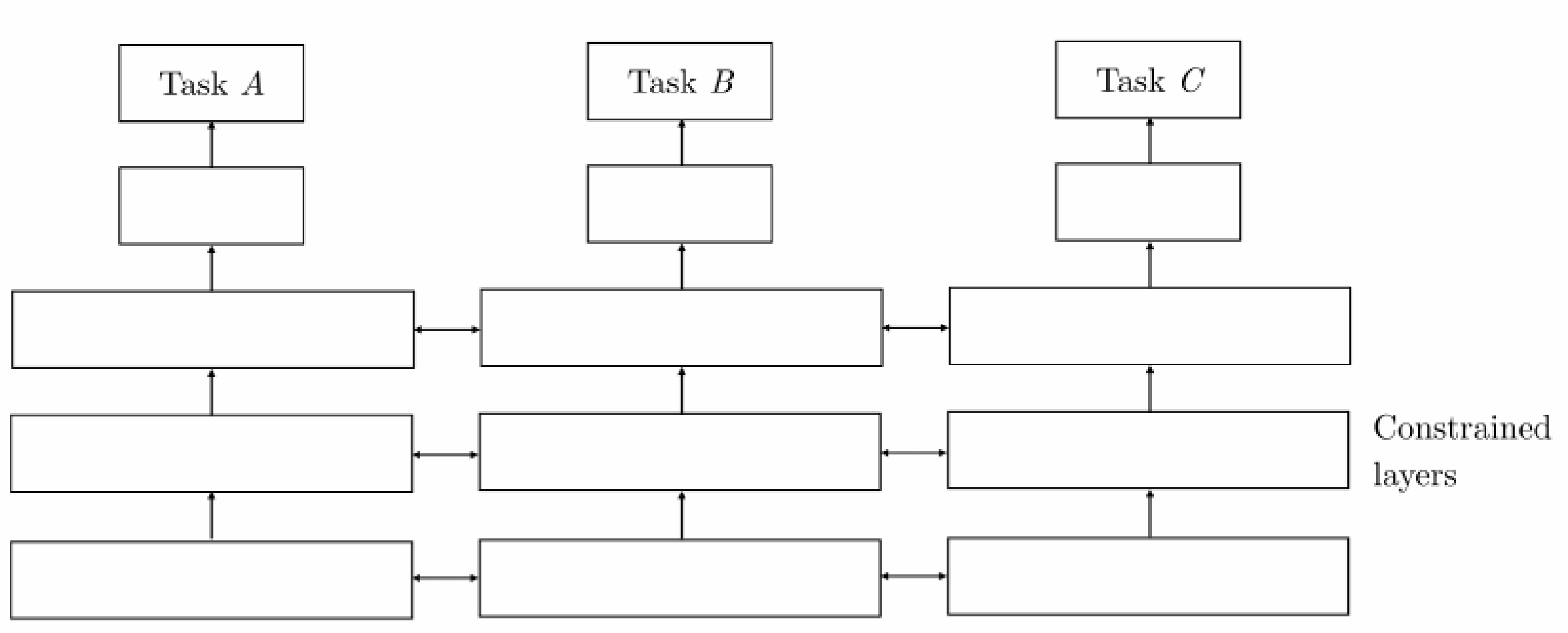

**Figure 3.5:** Soft parameter sharing in multitask learning [23]

The few reasons for the apparent functioning of multitask learning are:

- Latent data enhancement:
  As in multitask learning, the representation for more than one task is being learned simultaneously, the model also has to deal with different noise patterns dependent on each task and hence it helps in improving the overall representation of the model on the data.

- Representation preference:
  When learning for more than one task, the model tends to adopt the representation favored by all the task. This can help the model for the novel tasks as the representation space preferred by most of the tasks for higher performance would also be helpful for the novel task with its performance.

- Attention fixation:
  When a model is trying to learn on noisy or inadequate amount of data it could be difficult for the model to decide what features to focus on and what to ignore. MTL helps the model to focus on the important features by giving additional proof to help differentiate between pertinent and impertinent features.

### 3.2.2 Sequential Transfer Learning

It is a method of transfer learning in which the transfer of information for a deep learning model happens sequentially between tasks. First the model is trained on one task called source task in order to improve its performance on another task called target task. Due to this method of passing on information for the model from one task to another, it is also called as model transfer [25]. Sequential transfer learning is mostly advantageous when a) there is lack of data for both the tasks at a time; b) more data is available for the source task compared to the target task and c) there is a need for the model to get accustomed to more than one task.

#### 3.2.2.1 Phases in sequential transfer learning

There are basically two stages in sequential transfer learning:

1. Pretraining stage:
   It is the phase where the training of the model is done on the source task. It is generally costlier than the adaptation stage, however it has to be performed only once. As the task has to be performed only once it is very important to choose proper task that should be able to apprehend properties that can be advantageous for a good number of target tasks. The tasks that could be helpful for most NLP tasks are called universal tasks. The main factor that is used to discriminate between pretraining tasks is the source of supervision as an excellent pretraining task. It would provide a large amount of pretraining data, and raw unannotated data can be found effortlessly. The following three sources are arranged on the basis of difficulty of finding training data:
   - Classical supervision: This kind of supervision requires manually annotated training data and hence is the most challenging type of source of supervision to find.
   - Distant supervision: It basically tries to obtain sizeable chunk of noisy supervised data on the basis of domain expertise and heuristics. [26]
   - Without supervision: It is the most straightforward way as it just need raw data like the unannotated text.
2. Adaptation stage:
   In this stage, the transfer of knowledge happens from the trained model to the target tasks. Compared to the pretraining stage, this stage is more efficient and the work is limited to two major methods: a) feature extraction and b) fine-tuning, of the pretrained model to adapt it to the target task.
   - Feature extraction: In this stage, very much alike to the classical feature based approaches, the pretrained representations are used in the succeeding model after the model's weight are fixed [27]. For example, using pretrained word representations as an extra input to a model [28].

- Fine-tuning: This stage is mostly concerned with updating and productively using the pretrained representations for the succeeding task as initialization for the parameters of the model.

### 3.2.3 Cross-lingual learning

There are many ways of cross-lingual learning but this section is mostly focused on cross-lingual word embeddings. Cross-lingual word embedding basically provides a joint embedding space containing cross-lingual representations of words as shown in figure 3.6. Cross-lingual word embeddings provides word representations that can help us in figuring out meaning of a particular word in context of multiple languages and hence is considered crucial for cross-lingual transfer learning for low resources languages. The success of monolingual word embeddings coupled together with the development in multilingual benchmarks [30] [31] [32] and the increase in the acknowledgement of the digital language divide [33] has led to the surge in the research on the cross-lingual transfer learning and hence the cross-lingual word embeddings. Cross-lingual word embeddings are tempting for two reasons: a) it allows the inter-language comparison of the definition of a word which is important for tasks like cross-lingual information retrieval, machine translation, etc. and b) it helps in the interlanguage task of model transfer by accommodating a shared space of representation like that of between languages with variable amount of resources.
However the research in the cross-lingual learning is not motivated just by the success of word embeddings. Many researches have been done even before like learning from limited bilingual supervision, using seed lexica, document-aligned data or parallel data to learn word representations for cross-lingual settings.
The cross-lingual word embedding models are divided into three types on the virtue of data needed for the alignment across languages (i.e. whether the data in both the languages are aligned word by word, or sentence by sentence, or document by document):

- Word-level alignment:
  This approach basically relies on the pair of translations of words from cross-lingual dictionary building parallel word-aligned dataset [34] [35].

- Sentence-level alignment:
  This approach relies on parallel language corpus similar to machine translation.

- Document-level alignment:
  This approach relies heavily on the availability of parallel document aligned corpus which are essentially translations of one another. It is scarce to find this kind of parallel document aligned corpus as mostly the parallel documents are sentence aligned.

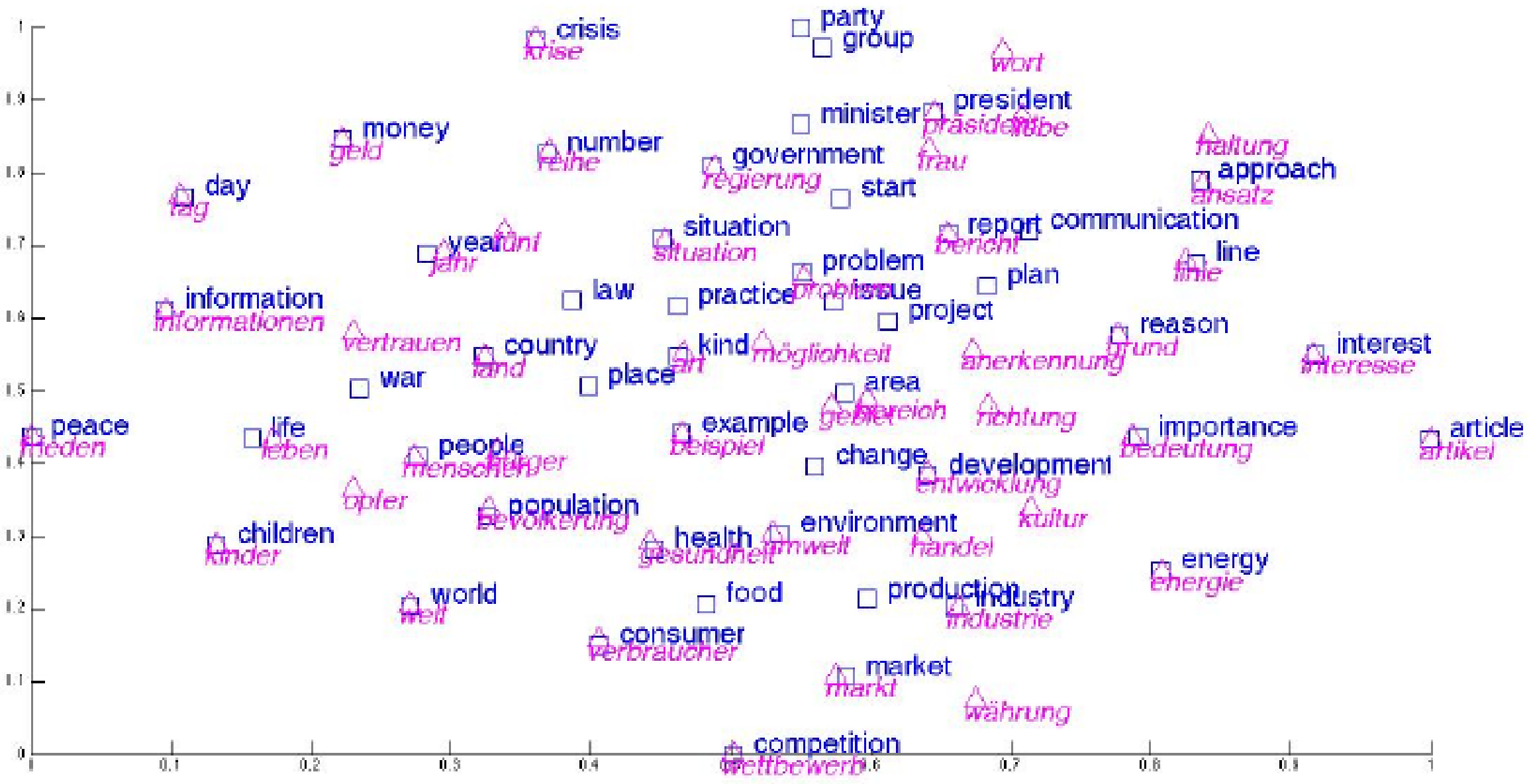

**Figure 3.6:** Example of cross-lingual embedding space [29] [23].

### 3.2.4 Domain adaptation

Domain adaptation is a method of transfer learning which is mostly concerned with adapting the model from a kind of training distribution to a different kind of test distribution. It is mostly focused on trying to learn representations that are more relevant to a specific target domain than learning representations that are helpful in general. Domain adaptation is considered mostly with unsupervised target setting with plentiful annotated data in the source domain and unannotated raw data in the target domain.

# 4 Hindi and Nepali Language

This chapter gives a short overview into the Hindi and Nepali languages and then it uses a statistical language similarity comparing websites to depict the similarity between both the languages.

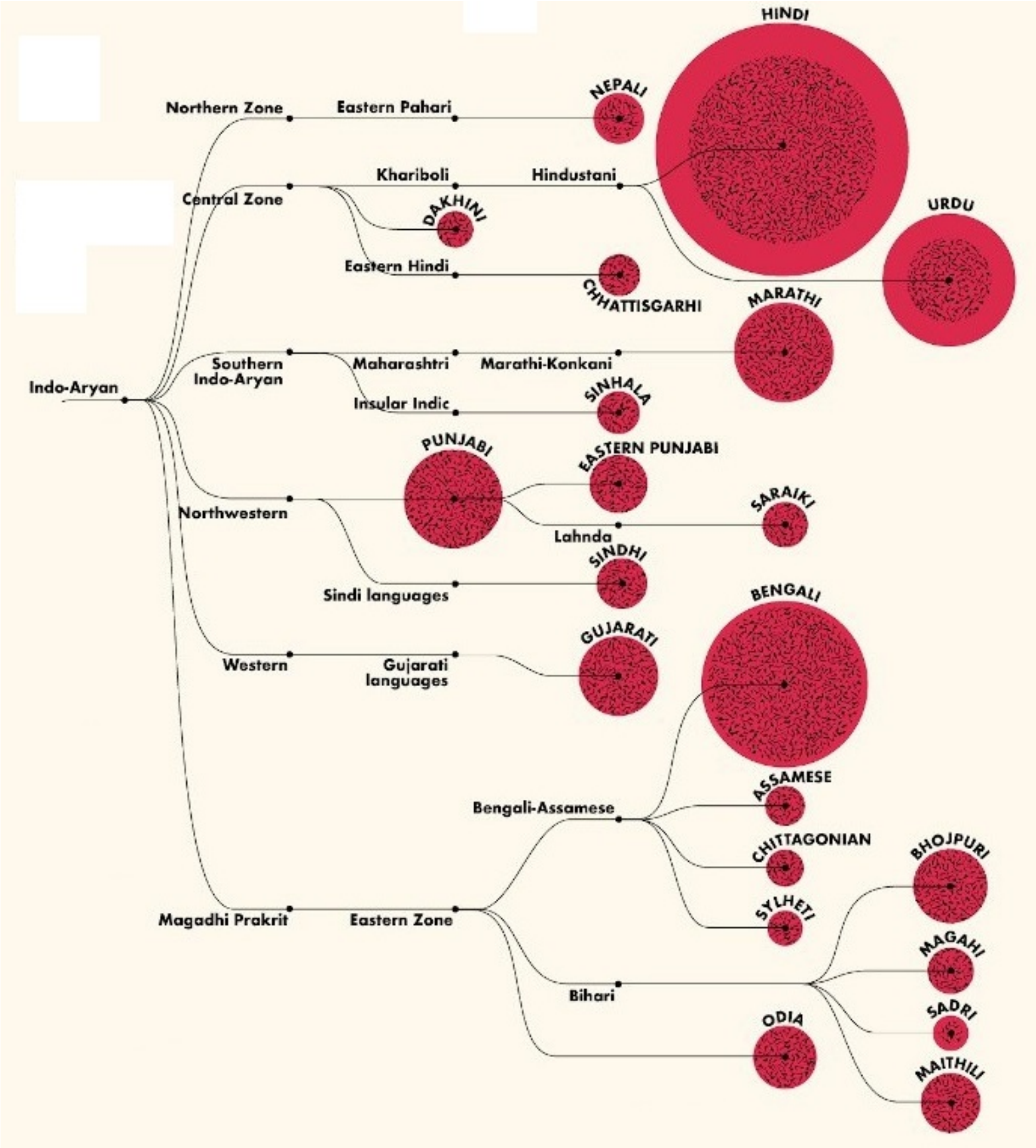

**Figure 4.1:** Indo-Aryan languages (see [36]).

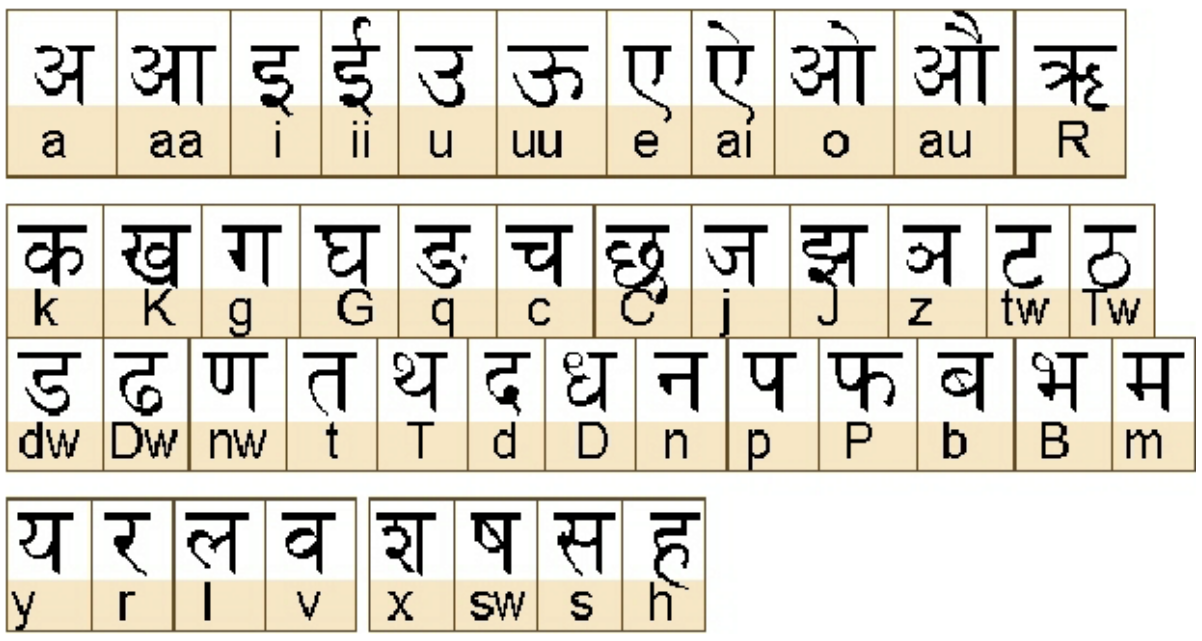

**Figure 4.2:** Devanagari script (see [37]).

Nepali and Hindi belong to the Indo-Aryan family of languages as shown in figure 4.1. Both the languages are derived from Sanskrit language and both use Devanagari script for writing as shown in figure 4.2 [38].
Hindi or Modern Standard Hindi is mostly spoken by people in the northern part of India called Hindi belt and is one of the official languages of India. It is the world's fourth most spoken language with 637.3 million total speakers [39]. [40]
Nepali is the official language of Nepal with 25 million total speakers worldwide. It is also spoken in by some as a mother tongue in Bhutan and India. [38]

According to a language similarity comparing website **elinguistics.net**, the genetic proximity between both these languages is *19.4*, where *0* refers to the closest languages and *100* refers to the most dissimilar language. The website uses statistical context to calculate the genetic proximity (for more information refer to: [41]).
Figure 4.3 shows the comparison of Hindi and Nepali language with some examples. The examples were generated randomly by the website mentioned above.
As it can be seen from both the examples and also from the genetic proximity score that the languages are quite similar. Therefore, Hindi and Nepali languages were chosen to explore transfer learning from one language to another.

| English | Hindi | Nepali | Comments | Points |
|---|---|---|---|---|
| **Death** | **-M-T-**<br>Maut (मौत) | **-M-R-T-**<br>Mrytyu (मृत्यु) | Exact consonant match -M-/-M-<br>Exact consonant match -T-/-T- | 66,67 |
| **Ear** | **-K-N-**<br>Kân (कान) | **-K-N-**<br>Kân (कान) | Exact consonant match -K-/-K-<br>Exact consonant match -N-/-N- | 100,00 |
| **Eye** | **-N-KH-**<br>Ankh (आंख) | **-N-KH-**<br>Ânkhâ (आँखा) | Exact consonant match -N-/-N-<br>Exact consonant match -KH-/-KH- | 100,00 |
| **Four** | **-CH-R-**<br>Châr (चार) | **-CH-R-**<br>Câr (चार) | Exact consonant match -CH-/-CH-<br>Exact consonant match -R-/-R- | 100,00 |
| **Hand** | **-H-TH-**<br>Hath (हाथ) | **-H-T-**<br>Hât (हात) | Exact consonant match -H-/-H-<br>Related consonant match -TH-/-T- | 89,12 |
| **I** | **-M-N-**<br>Mèn (मैं) | **-M-**<br>Ma | Exact consonant match -M-/-M- | 50,00 |
| **Name** | **-N-M-**<br>Nâm (नाम) | **-N-M-**<br>Nâm (नाम) | Exact consonant match -N-/-N-<br>Exact consonant match -M-/-M- | 100,00 |
| **Night** | **-R-T-**<br>Rât (रात) | **-R-T-**<br>Rât (राती) | Exact consonant match -R-/-R-<br>Exact consonant match -T-/-T- | 100,00 |
| **Nose** | **-N-K-**<br>Naak (नाक) | **-N-K-**<br>Nak (नाक) | Exact consonant match -N-/-N-<br>Exact consonant match -K-/-K- | 100,00 |
| **Sun** | **-S-R-J-**<br>Sûrya (सूर्य) | **-S-R-**<br>Suryâ | Exact consonant match -S-/-S-<br>Exact consonant match -R-/-R- | 66,67 |
| **Three** | **-T-N-**<br>Tīn (तीन) | **-T-N-**<br>Tin (तीन) | Exact consonant match -T-/-T-<br>Exact consonant match -N-/-N- | 100,00 |
| **Tongue** | **-ZH-B-**<br>Jibh (जीभ) | **-ZH-V-R-**<br>Jivro | Exact consonant match -ZH-/-ZH-<br>Related consonant match -B-/-V- | 62,21 |
| **Tooth** | **-D-N-T-**<br>Dant (दांत) | **-D-T-**<br>Dât (दाँत) | Exact consonant match -D-/-D-<br>Exact consonant match -T-/-T- | 66,67 |
| **Two** | **-D-**<br>Dô (दो) | **-D-**<br>Dui (दुइ) | Exact consonant match -D-/-D- | 100,00 |
| **Water** | **-P-N-**<br>Pânī (पानी) | **-P-N-**<br>Pâni (पानी) | Exact consonant match -P-/-P-<br>Exact consonant match -N-/-N- | 100,00 |

**Figure 4.3:** Comparison of Hindi and Nepali words similarity (see [41]).

# 5 Related Work

The chapter provides the state of the art in the task addressed in this thesis work. We look into the prior researches focused on the task of a part of speech tagging in general and then we narrow it down specifically to Hindi and Nepali languages with a focus on deep learning.

## 5.1 Part of Speech tagging

The research in the task of POS tagging in NLP has been undergoing right from the 1960s when the Brown English corpus was put forth at the Brown University by W.Nelson Francis and Henry Kučera which was POS tagged over the years. The first POS tagging was done by a program written by Greene and Rubin which achieved 70% accuracy. Then the hidden Markov models were started to be used for the task from the mid-1980s for the task. Currently, most of the machine learning algorithms like maximum entropy classifier, nearest-neighbor, and SVM can achieve accuracy above 95% on the various corpus.(see [42])
Due to the recent advances in deep learning techniques, there has also been great interest and research in using these techniques for the POS tagging. Plank et al. [43] uses Bi-LSTM with embeddings of words, and Unicode byte embeddings for the task. Vaswani et al. [44] uses a feed-forward neural network for the task and shows that encoding the information of complete sentences is important for better performance on the task. Yang et al. [45] use transfer learning for the task of POS tagging where they train the model on POS tags on a plentiful dataset and uses it to improve on the same task in different domains with a lesser amount of annotated data. They achieve the state of the art performance for the task. Ma et. al. [46] use the BLSTM-CNN-CRF architecture which was primarily developed for the task of Named Entity Recognition on the task of the POS tagging. The model provides comparable state-of-the-art performance[1].

## 5.2 Part of Speech tagging in Hindi and Nepali language

According to a survey done by Mehta et al. [47] on POS tagging in Indian languages, the work on Indian languages and in particular Hindi has not been so extensive given the lack of annotated corpus. Another reason for the lack of intensive work is the morphological richness and agglutinativeness of the language.
Initial work of Rule-based POS tagging on the Hindi language was done by Garg et.al [48] on 26,149 words and they tested the tagger on different domains of Hindi corpus like essays, news, and stories. The method was to first look in the corpus and tag those words which are found in the database and if the word is not found in the database then use the rule for tagging. Shrivastava et al. [49] uses a simple hidden Markov model for the task. Jha et al. [50] use a deep neural network to multitask on the morphological tags like POS tags, Gender, Number, etc. The basic ideas to try to optimize multiple loss functions simultaneously for each task in order to improve the effectiveness of the training.

[1] http://nlpprogress.com/english/part-of-speech$_{t}$*agging.html*

There has not been much work on the Nepali language when it comes to part-of-speech tagging as there has been a lack of properly annotated dataset for the language and also a lack of correct handwritten rules. Paul et al. [51] used a hidden markov model for the task as it just requires the information about the context of the language and hence could compensate for the lack of enough labeled dataset. Yajnik et al. [52] use an artificial neural network(i.e. feed-forward neural network) for the task.
The next chapter gives an overview of the deep learning model used to explore transfer learning between both languages.

# 6 Neural Network Architecture for Transfer Learning

There is a variety of neural network architectures available and it was really difficult to narrow down to one model for the implementation. However, for this particular task a novel architecture by Ma et al. [53] called, "End-to-end Sequence Labeling via Bi-directional LSTM-CNN's-CRF" was used.
The reasons for choosing this particular neural network architecture are:

- When we look into present-day researches, every other day we find papers being published claiming to have achieved better performance than the previous architectures but unfortunately, most of the time, it is so only on the carefully modeled datasets. These architectures fail to perform well in real-world scenarios. However, this architecture performs well in real-world scenarios on various tasks (see [54] [53]) .

- The architecture is mostly adopted by the industries given its performance and is mostly implemented by researchers for NLP tasks like Named Entity Recognition (see [55]).

- The architecture is not so resource-intensive as other latest architectures (see [56]).

- This architecture provides comparable state-of-the-art results compared to the latest architectures (see [54]).

## 6.1 CNN module

The Convolution Neural Network(CNN) has been really helpful when it comes to the task of image classification and object detection. The main advantage of using CNN is that it is very effective when it comes to extraction edges and tiny patterns in say, images, or in the sequence of phonemes in speech recognition. The main component of CNN is the convolution layer which is a variant of a fully connected layer used in a feed-forward neural network. Normally feed-forward network takes in a 1-dimensional vector and generates another 1-dimensional vector after matrix multiplication whereas convolution layer could take more than 1-dimensional inputs (like images with width, height and, channels). The main idea of the convolution layer is that the output neurons are connected to a small region of neurons in the previous layer to find important features which are position invariant in the input. The fundamental computation is shown in figure 6.1 which is to convolve or slide a window function by applying it to the input. The output indexes are obtained by convolving the values in the input with the values of the kernel element-wise by sliding the kernel and summing the value up.

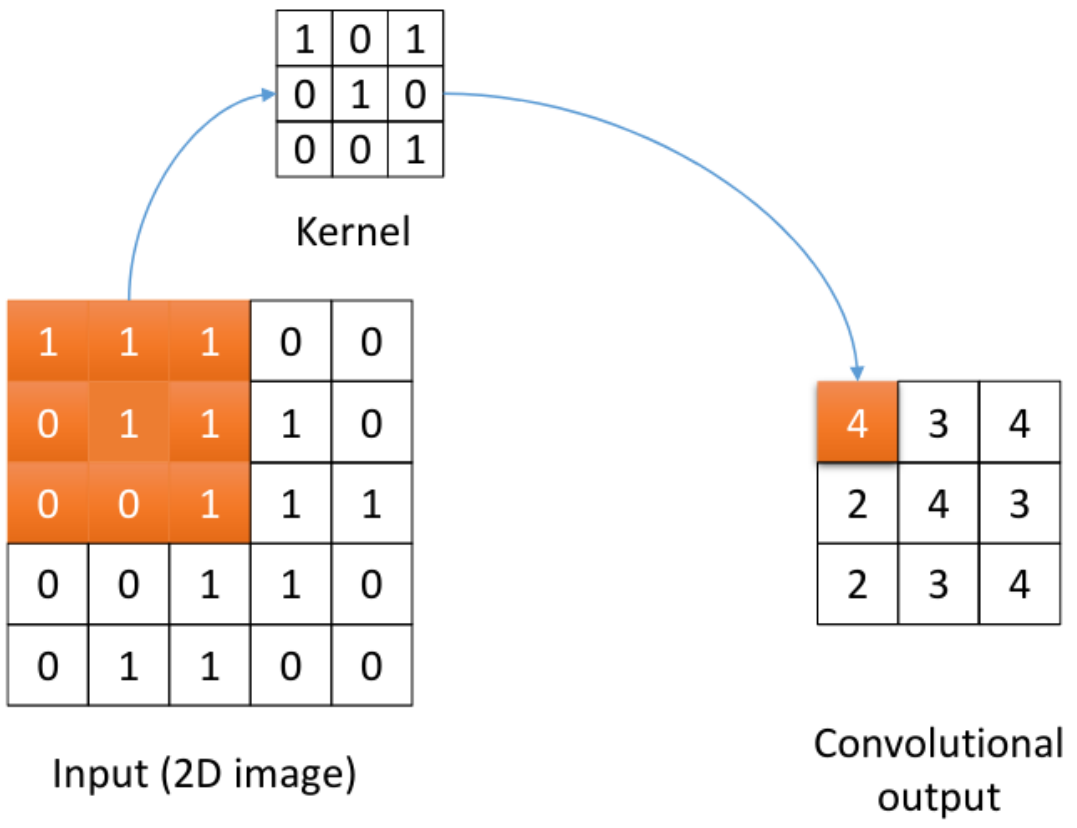

**Figure 6.1:** Illustration of Convolution [57]

In CNN, each input goes through a set of convolution layers which contain filters, fully connected layers, pooling layers, and finally the activation function. The example of it being used on an image is shown in figure 6.3. When an input image is fed into the neural network, it sees it as an array of pixels, and based on the image resolution it sees varying height, width, and dimensions. In the convolution layer, a feature map is produced which is a result of the convolution of the image matrix with the kernel matrix. The type of kernel matrix used decides which features are to be extracted. When the image size doesn't fit the kernel size, padding is used which for example with zero paddings, is it pads the part of the picture with zeros so that it could fit properly. Then the pooling layer is used to discard the unwanted information when the image is too large. In max-pooling as shown in figure 6.2, the largest element from the feature map is selected and rest are discarded. The output of max-pooling layer is fed to the fully connected layer where the feature map matrix is flattened into vector and fed as input.

It combines all these features to finally create a model. And finally, the output is fed to an activation function like softmax or sigmoid to classify the object in the image.

The convolution neural networks help in extracting local features of the data more efficiently. CNN was used by Chiu and Nicholas et al. [59] for the task of extracting morphological information of words or characters which was then further used for the task of named entity recognition. The specific neural network architecture of the CNN used for the task is shown in the figure 6.4. Each character of the word is transformed into character embeddings by the help of look up table. A lookup table basically contains numerical values which could be used to substitute and represent each character of a given word. The character embeddings passes through a dropout layer before being fed as input to the CNN. And then finally, max pooling is used to filter out important information and discard not so useful information.

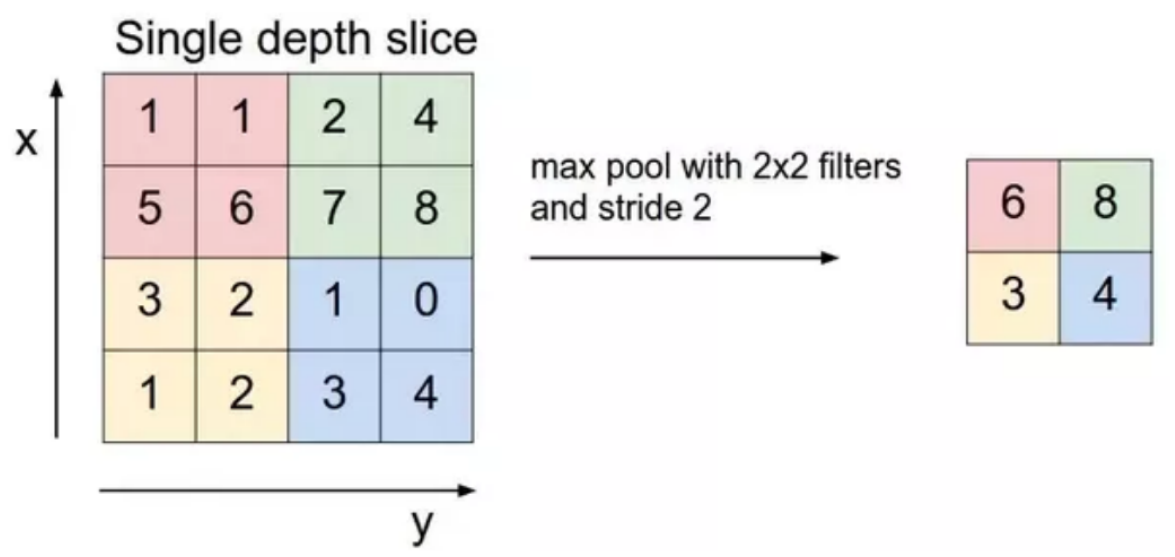

**Figure 6.2:** Max Pooling [57]

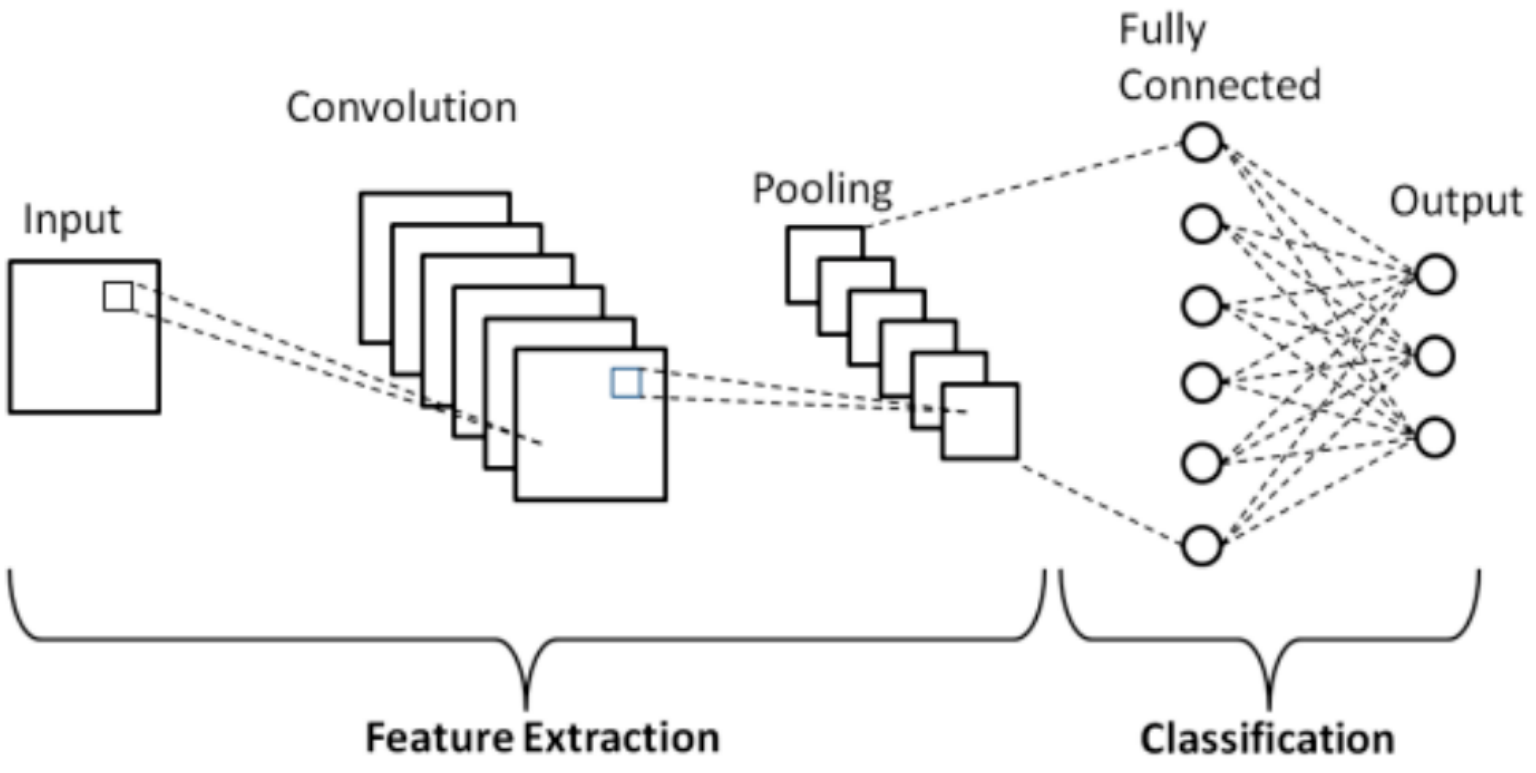

**Figure 6.3:** Architecture of CNN [58]

## 6.2 BLSTM Module

Bi-directional Long Short Term Memory consists of two Long Short Term Memory(LSTM) blocks. LSTM is a typical version of the Recurrent Neural Network(RNN). Recurrent Neural Network is a type of neural network which is used to develop sequential networks where data are dependent on each other and also the sequence of data matters (for detailed information on RNN, please refer to: [60]).

Sequential data can be best simulated with the help of a Recurrent neural network(RNN). However, as the traditional RNNs couldn't take care of the property of sequential data like "long-range dependence" and leads to gradient disappearance or gradient explosion [62]. This problem can be taken care of by the Long Short-Term Network(LSTM) proposed by Hochreiter et al. [63].
LSTMs were put forth by Hochreiter Schmidhuber. It is used to retain information from previous inputs and it basically helps in solving the NLP problems like vanishing gradient and long term dependency. It processes data in the forward direction and has the capacity to add or omit data that are controlled by gates.
It has three gates, i.e. input gate, output gate, and forget gate which controls the passage of information. The internal architecture of a LSTM unit is shown in figure 6.5. It uses tanh the activation function for the purpose of cell state activation and the node output is controlled by a sigmoid activation function. The advantage of LSTM is that it contains a gate structure which helps in controlling the model to retain contextual information selectively. The LSTM unit can be formalized with the following equations at time t:

$$\begin{aligned} i_t &= \sigma\left(W_i h_{t-l} + U_i x_t + b_i\right) \\ f_t &= \sigma\left(W_f h_{t-l} + U_f x_t + b_f\right) \\ \tilde{c}_t &= \tanh\left(W_c h_{t-l} + U_c x_t + b_c\right) \\ c_t &= f_t \otimes c_{t-l} + i_t \otimes \tilde{c}_t \\ o_t &= \sigma\left(W_o h_{t-l} + U_o x_t + b_o\right) \\ h_t &= o_t \otimes \tanh\left(c_t\right) \end{aligned} \tag{6.1}$$

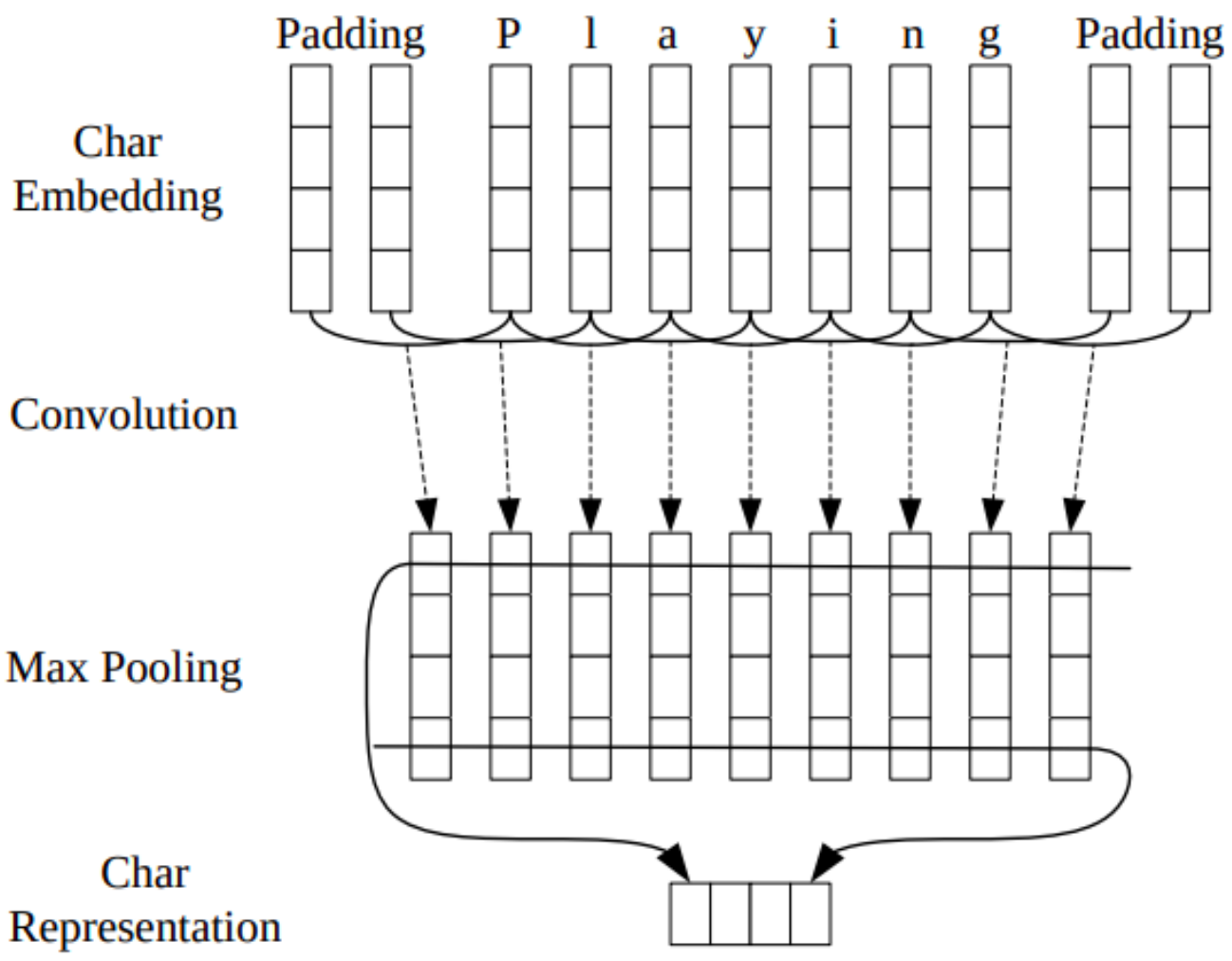

**Figure 6.4:** Architecture of CNN module (see [53]).

Where, $\sigma$ represents the sigmoid activation function; $\otimes$ represents point-wise multiplication which means the multiplication of elements with the same positions in both vectors; tanh represents the hyperbolic tangent function; $x_t$ represents the input vector at a given time $t$ ; $h_t$ represents the hidden state of the LSTM at time $t$ ; $i_t, f_t, o_t$ is the input gate, forget gate and the output gate $t$; the weight matrix of the input vector $x_t$ is represented by $U$ for each gate; $W$ and $b$ represents the hidden state's weights and biases at each gate; $\tilde{c}_f$ is the intermediate state gained according to the input at a given time $t$ ; used to update the current state; $c_t$ is the state at time $t$ [53].

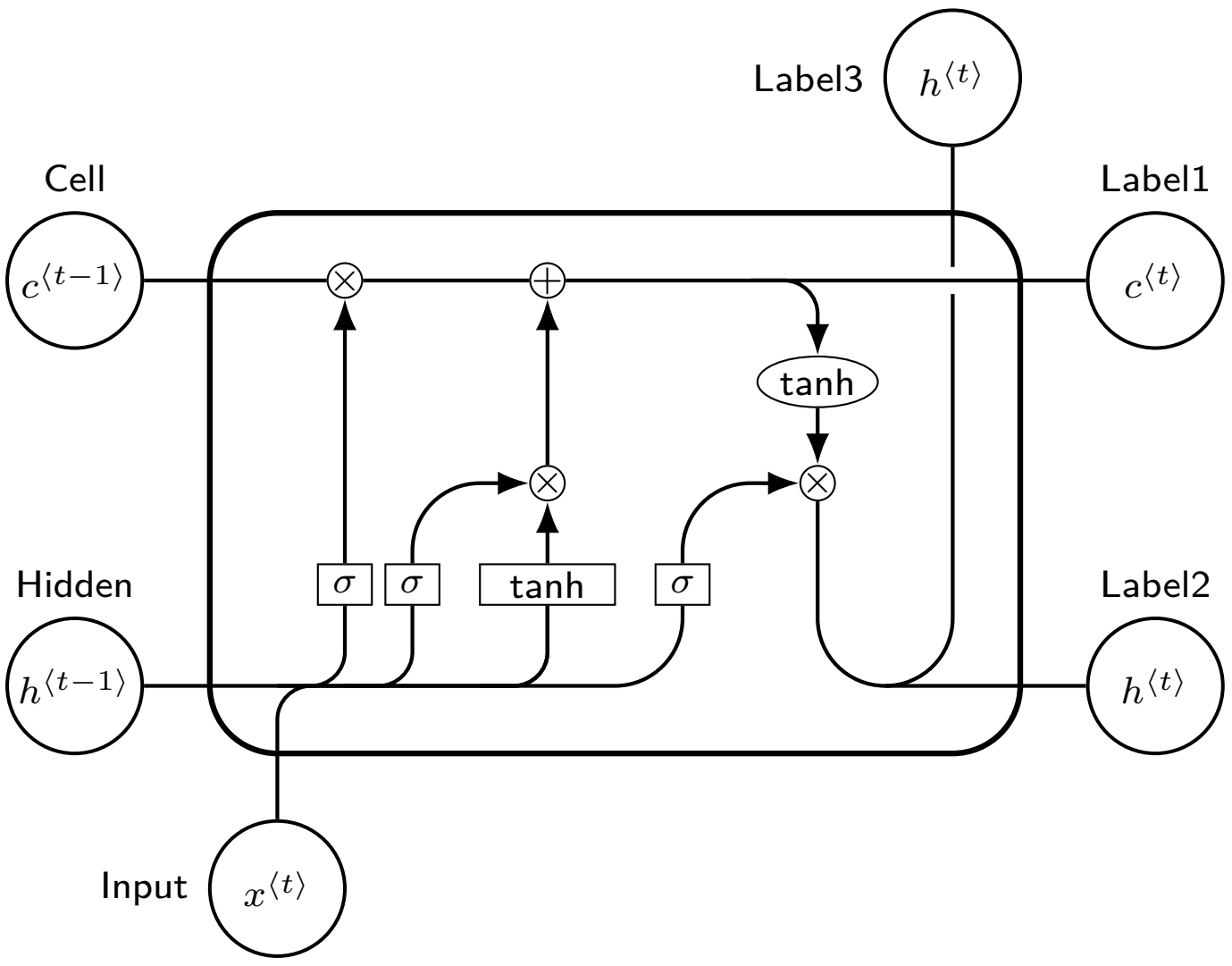

**Figure 6.5:** Internal architecture of LSTM [61].

The problem with LSTMs is that they can only deal with data in the forward sequence. Hence, in this work, the bidirectional LSTM(BLSTM) is used for better use of contextual information where one LSTM looks at the sequence of words in a sentence in the forward direction and others in the backward direction. The main idea is to extract two different hidden states by the forward and backward LSTM and which is used to represent future and past in the sequence which is further concatenated together. This concatenated output is passed on to the next module.

## 6.3 CRF Module

Conditional Random Field can be used to extract best label sequence given the relationship between adjacent labels. For example, in the present task of POS tagging, normally a noun cannot be followed by another noun and there is a high probability that it be followed immediately by verb. Hence, it is helpful to consider the tags given to the surrounding words to predict the tag of the considered word. The output of the previous BLSTM layer is fed to the CRF layer to get final tags for the words in the sentence. $X = \{x_1, x_2, \cdots, x_n\}$ is taken as the vector representation of the words in the sentence and the corresponding label sequence is given by $Y = \{y_{l,}y_2, \cdots, y_n\}$. Hence, the CRF probability model can be taken as the probability $p(y|x; W, b)$ by the corresponding label represented by $y$ under the given variable condition represented by $x$. It can be depicted by the following formula:

$$p(y|x; W, b) = \frac{\prod_{i=1}^{n} \Psi\left(y_{t-1}, y_t, x\right)}{\sum_{y_i)(z)} \prod_{i=1}^{n} \Psi_t\left(y_{t-l}, y_t, x\right)} \tag{6.2}$$

Where a potential function is represented by $\Psi_t(y, y, x) = \exp\left(W^T_{y',y}x_t + b_{y:y}\right)$, bias of the tag pair $(y', y)$, and the weight vector is represented by $b_{y:y}$ and $W^T_{t',y}$ respectively. The formula to calculate the log-likelihood function is given by:

$$L(W, b) = p(y|x; W, b) \tag{6.3}$$

Maximum likelihood estimation is used to get the optimal solution of best possible sequence of predicted tags for a given input sentence.

## 6.4 BLSTM-CNN-CRF

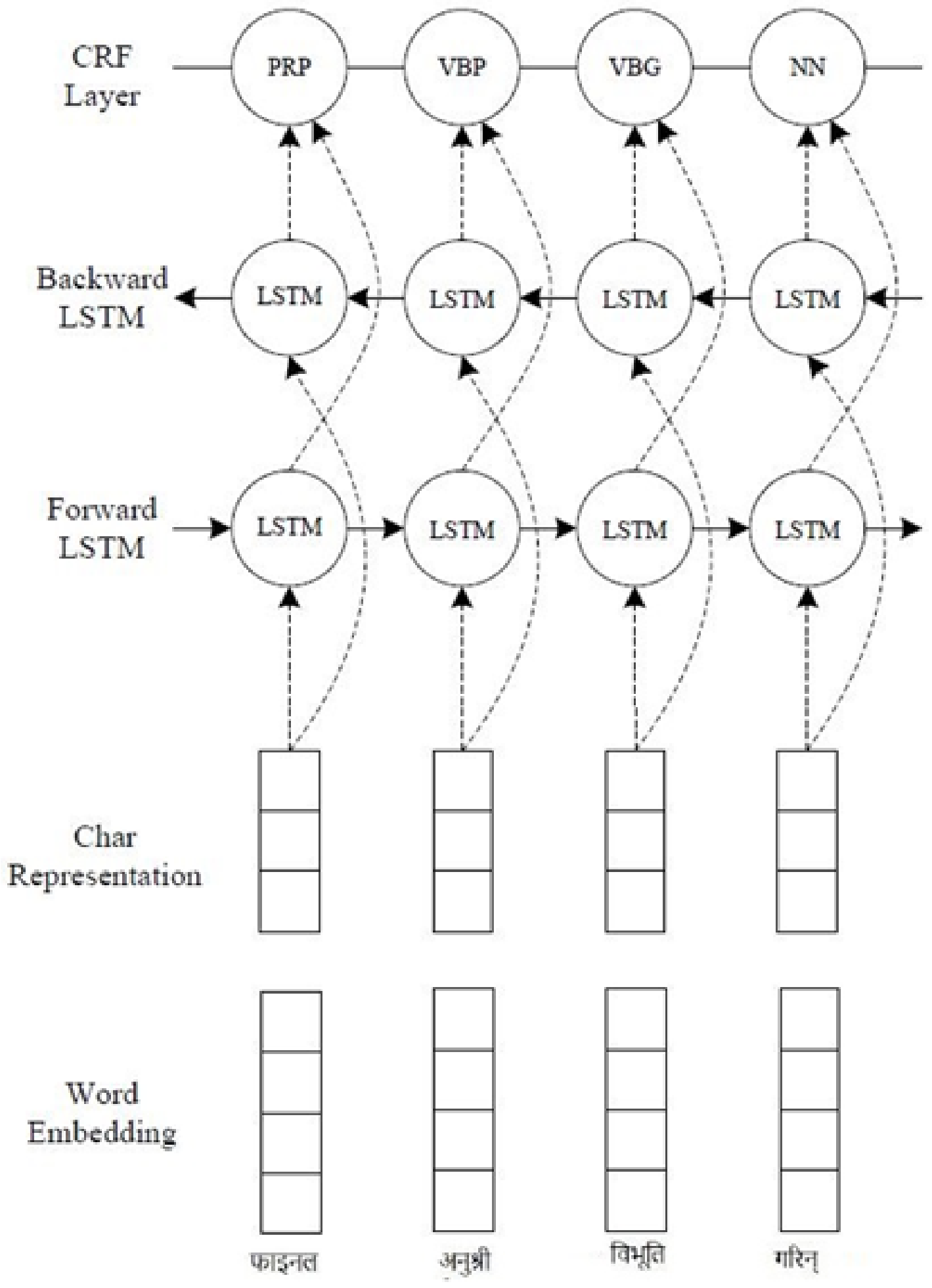

**Figure 6.6:** Architecture diagram of BLSTM-CNN-CRF (see [53]).

Figure 6.6 shows the framework of the neural network architecture put forth by Ma et. al [53]) for end to end sequence labeling tasks in natural langauge processing. This approach proposes the use of end-to-end sequence labeling architecture combining Bidirectional Long Short-Term Memory (BLSTM), Convolution Neural Network (CNN), and Conditional Random Field (CRF).
CNN is used to extract the character level information with input being fed as character embeddings of the input words. CNN is used to extract the character level morphological information like the suffix or prefix of a word and encodes it into neural representation. Then the character embeddings is concatenated with the word embeddings which is further fed into the BLSTM that models the contextual information. BLSTM basically contains two LSTMs where one gets the concatenated embeddings of the sentence in the forward direction and the other gets it in the backward direction as it is beneficial for the model to have future and past contexts. And the output of both the LSTMs concatenated together. The dotted line as can be seen in figure 6.6 represents application of dropout layers which has been seen to improve the performance of the model significantly. As it can be seen, it has been applied to both input and output vectors of the BLSTM. Dropout is a method used to prevent overfitting of the model on the training data by avoiding the updating of weights on a few hidden nodes.
And finally, the tag for each word is generated by the sequential CRF layer which jointly decodes the labels for the whole sentence fed into the model.
The loss function used to calculate the gradient which are further used to update the weights of the neural network is Sparse Categorical Cross-entropy(SCC) given by equation 6.4 [64]. Where $S$ represents samples, $C$ represents Classes and $s \in c$ - sample belongs to class $c$.

$$-\frac{1}{N}\sum_{s\in S}\sum_{c\in C} 1_{s\in c}\log p(s \in c) \tag{6.4}$$

SCC is used for multi-class classification where it compares the true and predicted tags and then calculates the loss. This loss function was used because the predictions from our model are from multiple classes and it represents each category is represented by an index instead of one-hot encoding which save the memory. [53].

# 7 Implementation

The previous chapter dealt with the concepts on the basis of which this thesis work is conducted. This chapter deals with the practical steps that need to be undertaken to complete this work.

## 7.1 Preliminary

Before the implementation of this work, there is a need to fulfilling a few prerequisites. This section deals with these requirements needed for the successful implementation of this work. This section deals with the information about the datasets required to in all the experiments.

### 7.1.1 Dataset sources

The POS tagged the Hindi dataset from Linguistic Data Consortium(LDC) [65] and the publicly available POS tagged Nepali corpus from the Center for Language Engineering(CLE) [66] in Pakistan was used for the purpose of transfer learning in this thesis. For the purpose of training word embeddings for both the Nepali and Hindi languages, the respective language corpus was used from Wortschatz at the University of Leipzig [67].

#### 7.1.1.1 Linguistic Data Consortium

Established in 1992, Linguistic Data Consortium(LDC) is an open organization created to focus on solving the problem of data shortage faced in the research and development of language technology. It is based in the School of Arts and Sciences at the University of Pennsylvania. [65]
The catalog number LDC2010T24 with ISBN 1-58563-571-5 was used for this work. This corpus has been developed by Microsoft Research India(MSR) in collaboration with linguistics and computer scientists from Anna University, Indian Institute of Technology(Bombay), Delhi University, Jawaharlal Nehru University, and Tamil University. It was set up under the *Indian Language Part-of-Speech Tagset(IL-POST)* project in order to assist the linguistic research in Indian languages. The IL-POST implements a common tagset framework that provides reusability, flexibility, and cross-lingual compatibility across the Indian Languages. The corpus is built with manually POS tagged 98,450 words(4859 sentences). The text was gathered from Microsoft Hindi Research Corpus. [65]

#### 7.1.1.2 Center for Language Engineering, Pakistan

Center for Language Engineering(CLE) is a research organization funded by the Pakistan government that conducts research and development on Asian languages and specifically on languages in Pakistan. The Nepali dataset used for the purpose of this thesis work has been created in parallel to Urdu and released by the Center of Research in the Urdu Language Processing(CRULP). The dataset is created in parallel to 100,000 English words from PENN Treebank corpus which is accessible at Linguistic Data Consortium. The creation of this dataset was assisted by the Japanese Language Resource Association(GSK) and the Canadian International Development Research Center(IDRC). [66]

#### 7.1.1.3 Wortschatz, University of Leipzig

Wortschatz is a multi-language corpora collection project created by the *Natural Language Processing Group* of the Institute of Computer Science at Leipzig University. It contains mono-lingual corpus for more than 250 languages in the same format and from the comparable sources. The texts are collected either from newspapers or randomly from the web. All the files are in plain text format in sizes ranging from 10,000 sentences to 1 million sentences. [67] [68]
The size, source and year of the building of the corpus used to train word embeddings are given in table 7.1.

**Table 7.1:** Word embedding corpus

| | Source | Size | Year |
|---|---|---|---|
| Hindi | Web | 300K | 2015 |
| Nepali | Web | 300K | 2015 |

## 7.2 Word Embedding Generation

The word embeddings were generated by using the fastText library created by Facebook [19] [20] [69]. The model was first trained on the Hindi dataset of 300K sentences as given in the previous section. The model was trained on GPU provided by Google Colab. Similarly, the new model was again trained on the Nepali dataset of 300K sentences to generate Word embeddings. And finally, the Nepali 300K sentences were put together with 300K Hindi sentences, and then the fastText model was trained on the combined dataset to generate combined word embeddings.

## 7.3 Unsupervised Cross-lingual Mapping of Word Embeddings

The method used for the cross-lingual word mapping of monolingual word embeddings of Hindi and Nepali languages is derived from the paper published by Artetxe et. al [70]. Till now other methods of mapping monolingual embeddings to a shared space by adversarial training have been failing in the real world scenarios. They propose an unsupervised method that exploits the structural similarities in the word embeddings of two different languages and additionally a self-learning algorithm that betters the solution iteratively. They don't require a seed dictionary for the initial mapping. The basic idea underlying this implementation is that if we consider the similarity matrix of all the words in a vocabulary, every word has similarity values of different distribution and they exploit and use this fact that two similar words in two different languages will tend to have similar distribution for the induction of the initial set of word pairs. Further, they use a self-learning method which starts with poorer solutions initially and improves on it iteratively. This method is completely unsupervised and could be also used in realistic settings. According to the authors, this method produces better results than even the previous supervised methods.
The proposed method is that let's assume $X$ and $Z$ be the word embedding matrices in two different languages and their respective $i$ th row $X_{i*}$ and $Z_{i*}$ represents embeddings of the $i$ th word in the vocabularies. The goal is to be able to learn the transformation matrices $W_X$ and $W_Z$ which brings the mapped embeddings $XW_X$ and $ZW_Z$ in the same cross-lingual space. And further, they try to build a bilingual dictionary between both the languages which can be encoded as a matrix $D$ where $D_{ij} = 1$, considering $j$ th word is a translation of $i$ th word of target and source language respectively.
The method proceeds with four steps wherein pre-processing phase the normalization of embedding takes place and then it goes on to the fully unsupervised initialization phase where the initial solution is created and then it moves on to the third phase of robust self-learning procedure where the solution is iteratively improved and finally in the refinement phase the final resultant mapping is further improved by symmetric re-weighting. (for more information on this approach refer to: [70])

## 7.4 Deep Learning Architecture

The basic architecture of the model is the same as explained in chapter 6. This section will deal with the implementation of POS tagging in the Nepali language and Hindi language. Then it will give an overview of the implementation of multitask learning in Hindi language and then finally it will explain the architecture used to implement transfer learning between Nepali and Hindi language for the task of POS tagging.

### 7.4.1 POS tagging Hindi and Nepali Language

Figure 7.1 depicts the flow diagram of the implementation of POS tagging for both Hindi and Nepali language datasets. As both the implementations uses the same architecture, only the implementation of Hindi has been explained. For a Hindi sentence which is fed to the architecture, character and subword level representation is fed to CNN. The words are padded before being fed to the CNN to maintain uniform length. The output of CNN is then concatenated with the pre-trained word embedding vector. Then this combination is fed as a whole to the BLSTM. And finally, the concatenated output of both LSTMs in BLSTM is fed to the CRF layer which decodes the POS tags for the entire sentence. While training, the decoded POS tags are compared with the actual tags, and the weight in the model is adjusted with the help of optimizer to be able to generate more accurate tags.

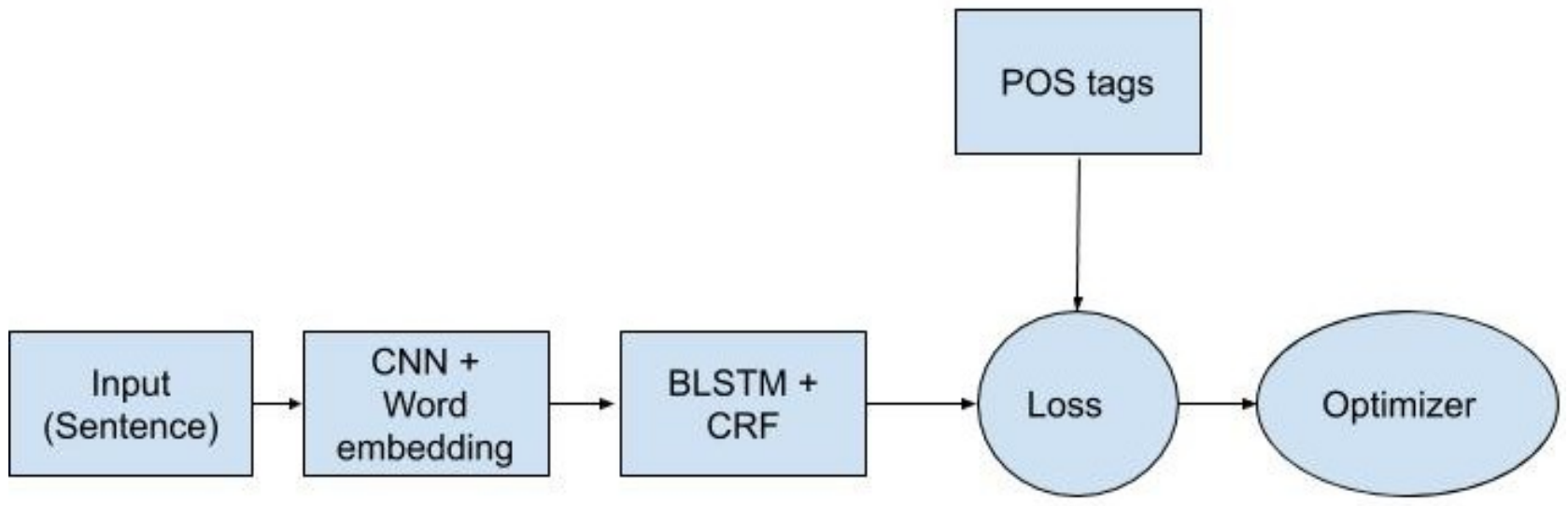

**Figure 7.1:** Workflow of implementation of POS tagging.

### 7.4.2 Multitask learning in Hindi language

The general step by step procedure of the implementation is shown in figure 7.2. The multi-task learning architecture implemented here is of hard parameter sharing where the initial layers of deep neural network is shared for all three tasks and then later on the layers are separated for each specific tasks. For each sentence fed to the architecture, as explained in the previous section, each character is transformed into character embeddings with the help of the lookup table. Then it is fed to the shared BLSTM network. Then the output of the shared BLSTM network is fed to the separate BLSTMs dedicated to the tasks of singular/plural tagging, POS tagging, and gender tagging. Then the output of these dedicated BLSTMs is fed the dedicated CRF layers which decode the tags for each task. While training, the generated tags are compared with actual tags, and optimizers are used to adjust the weight accordingly.

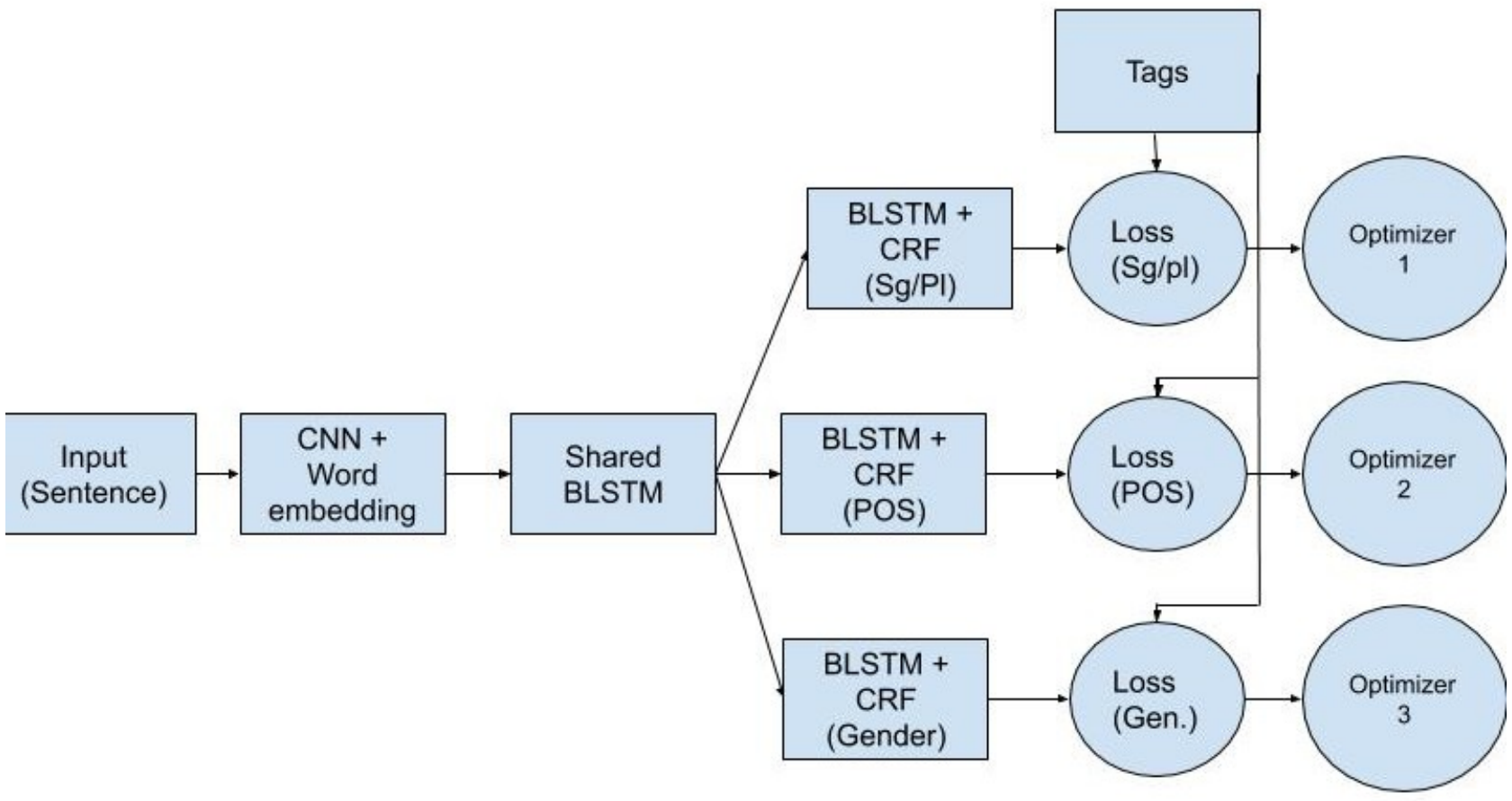

**Figure 7.2:** Workflow of implementation of Multitask learning

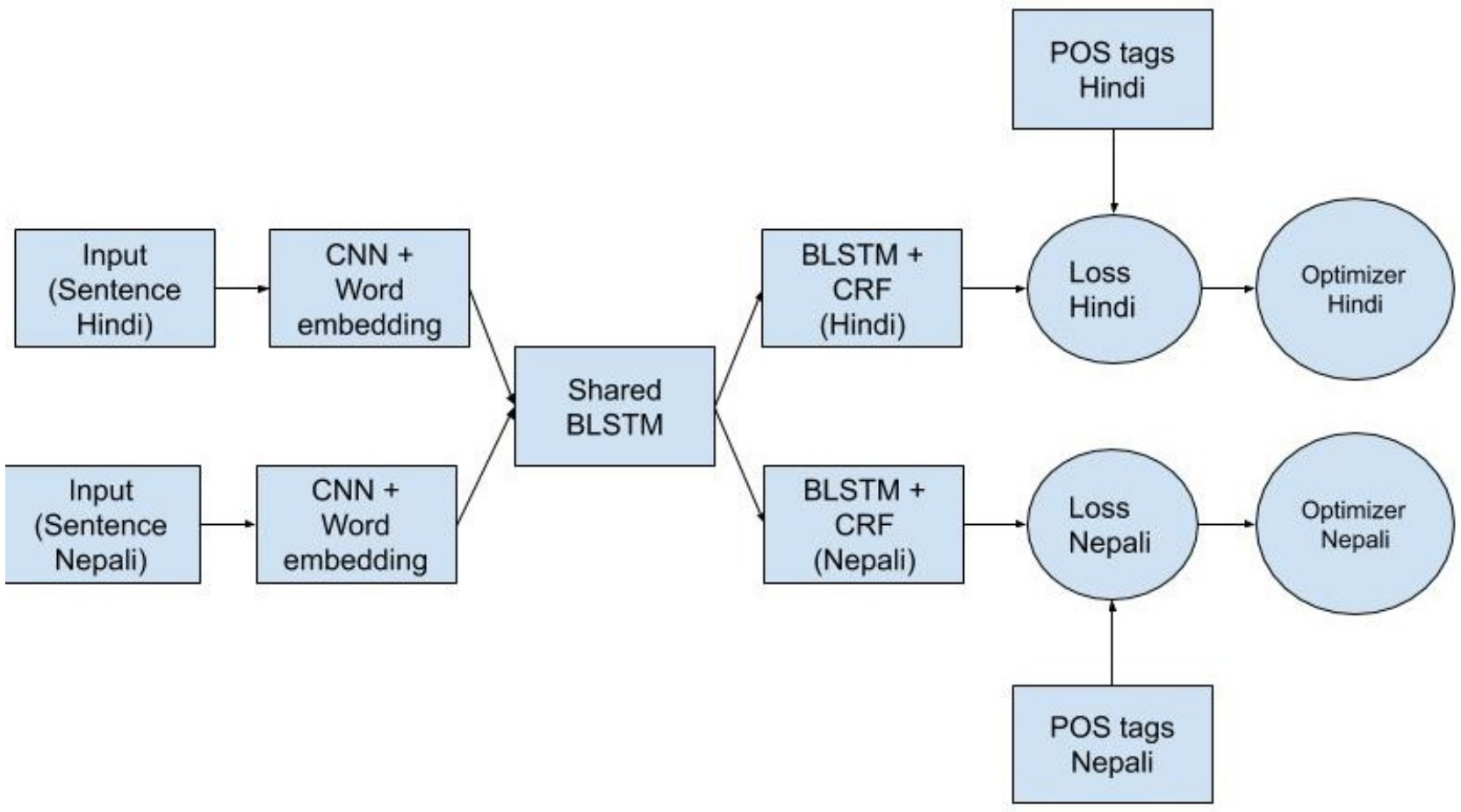

**Figure 7.3:** Workflow of implementation of Transfer learning

### 7.4.3 Transfer Learning between Hindi and Nepali

This section explains the architecture which was used in the implementation of transfer learning between Hindi and Nepali language. The transfer learning architecture implemented here is Cross-lingual learning where a model tries to solve similar task in two languages simultaneously. The architecture contains separate components for each task except for the shared BLSTM layer as seen in figure 7.3.
Hindi and Nepali sentences are fed to the respective CNNs. As both the language uses the same script, the same character embeddings are used. Then the output of the CNN is concatenated with word embeddings of individual language. It is fed together to the shared BLSTM and then the output of shared BLSTM is then fed to the individual language BLSTMs which is then further fed to the dedicated CRF layers. The CRF layers decode the POS tags for the entire sentence. While training, the loss is calculated by comparing the outputs of each task-oriented CRF with actual tags, and optimizers are used to reduce the loss. The training is done sequentially one sentence at a time for each language.

# 8 Experimental Study and Performance Evaluation

This chapter provides a detailed overview of all the experiments conducted in this thesis. It contains experiments on POS tagging in Hindi language, POS tagging in Nepali language, multitasking in the Hindi language, and transfer learning in between both the languages.

## 8.1 Experimental Setup

This section deals with the prerequisite conditions for the successful implementation and execution of all the experiments. It reports hardware used for the implementation, details about the software and finally reports the metrics used for the measure of the performance of the different approaches.

### 8.1.1 Hardware

PlanetAI GmbH supported me in the work for my master thesis by providing me with a Dell laptop to conduct any kind of experiments. The laptop contained a Intel HD Graphics 4600 processor. Furthermore, it has an Intel Core i5-4310U central processing unit. The CPU dual-core and has a base frequency of 2.6GHz. The RAM size of the laptop is 8GB. [71]

### 8.1.2 Software

The software and libraries were used for the implementation of this project.

- **Tensorflow**
  Tensorflow is an open-source software mathematical library developed by Google Brain for high-performance numerical computations. It uses C++ for the accomplishment of the applications and Python to setup applications with the framework which provides a convenient Application Programming Interface (API). It provides a high degree of flexibility with the implementation of computation irrespective of the platforms like Central processing units, graphics processing units or Tensor Processing Units. It can be accessed from all the most used operating systems like Windows, macOS, Linux and also on mobile computing platforms like iOS and

Android. [72]

- **Google Colaboratory**
  Google collaboratory was used for the purpose of vector mapping. It is a Jupyter notebook based research tool that can be used for machine learning related education and research. It can be accessed by major internet browsers like Chrome and Firefox. It supports Python with a virtual machine is dedicated to a particular google account for code execution which has a maximum lifetime as enforced by the system and which is recycled when machine stays idle for some time. Colaboratory is basically for collective use and hence it may interrupt long-running background computations on GPU. But this can be avoided by using a local runtime which supports continuous or long-running computations. [73]

- **Python**
  It is a high-level interpreted programming language for general-purpose created in 1991 by Guido van Rossum with code readability as its core concept. It has a very powerful language because of the variety of libraries available. [74]

## 8.2 Metrics

In this subsection, we talk about the accuracy metrics which was used for the evaluation of the model. Accuracy is one of the most intuitive metrics used to measure the performance of the model. It basically means the number of times the model got the correct predictions given the total number of predictions. The formal definition of accuracy is:

$$\text{Accuracy} = \frac{\text{Number of correct predictions}}{\text{Total number of predictions}} \tag{8.1}$$

For the models involving binaries(i.e. True and False), accuracy is calculated by the following formula:

$$\text{Accuracy} = \frac{TP + TN}{TP + TN + FP + FN} \tag{8.2}$$

Where $TP$ = True Positives, $TN$ = True Negatives, $FP$ = False Positives, and $FN$ = False Negatives.

## 8.3 Dataset

The count of the total words and sentences in test, train and development dataset of both Nepali and English dataset is given below.

- Hindi: 4252 sentences which were divided into train, test and development with 8:1:1 ratio as shown in table 8.1.

| Hindi | Train | test | dev |
|---|---|---|---|
| Tokens | 81564 | 10151 | 10130 |
| Sentences | 3551 | 399 | 300 |

**Table 8.1:** Hindi dataset

- Nepali: 4245 sentences which were divided into train, test and development with 8:1:1 ratio as shown in table 8.2.

| Nepali | Train | test | dev |
|---|---|---|---|
| Tokens | 71466 | 8918 | 8916 |
| Sentences | 3347 | 452 | 444 |

**Table 8.2:** Nepali dataset

The tags for the task of Part-Of-Speech tagging are given below:

1. Noun
    - *Common Noun (NN)*
    - *Proper Noun (NNP)*
2. Verbs
    - *Finite Verb (VBF)*
    - *Auxiliary Verb (VBX)*
    - *Verb infinitive (VBI)*
    - *Prospective participle verb (VBNE)*
    - *Aspectual participle verb (VBKO)*
    - *Other participle verb (VBO)*
3. Pronoun
    - *Personal Pronoun (PP)*
    - *Possessive Pronoun (PP$)*
    - *Reflexive Pronoun (PPR)*
    - *Marked Demonstrative (DM)*
    - *Unmarked Demonstrative(DUM)*
4. Adjective
    - *Normal/unmarked Adjective(JJ)*
    - *Marked Adjective (JJM)*
    - *Degree Adjective (JJD)*
5. Adverb
    - *Manner Adverb (RBM)*
    - *Other Adverb (RBO)*
    - *Location Adverb (ALC)*
6. Intensifier (INTF)
7. Postpositions (POP)

8. Conjuction
    - *Coordinating Conjunction (CC)*
    - *Subordinating conjunction (CS)*
9. Interjection(UH)
10. Number
    - *Cardinal Number (CD)*
    - *Ordinal Number (OD)*
11. Plural marker (HRU)
12. Question word (QW)
13. Classifier (CL)
14. Particle (RP)
15. Determiner (DT)
16. Unknown word (UNW)
17. Foreign word (FW)
18. Punctuation
    - *Sentence Final (YF)*
    - *Sentence Medieval (YM)*
    - *Quotation (YQ)*
    - *Brackets (YB)*
19. Abbreviation (FB)
20. Header list(ALPH)
21. Symbol (SYM)

## 8.4 Embeddings used

- Fasttext word embeddings for Hindi and Nepali:
  These embeddings were trained by using Fasttext[1] skip-gram model on the corpus obtained from the University of Leipzig [67]. The Hindi language was trained on 1M words dataset and Nepali has deliberately trained on just 300k words dataset.
- Vector aligned Fasttext word embeddings:
  The above Hindi and Nepali embedding vectors were cross-lingually words mapped using the unsupervised techniques put forth by Artetxe et. al. (for further information on this technique refer to: [75]).
- Nepali-Hindi word embedding together:
  In this part, a fasttext word embedding model was trained on Hindi and Nepali corpus together from the University of Leipzig to generate a common word embedding for both the languages.

All the above word embeddings were used to train the model for the following:

- Nepali POS tagging
- Hindi POS tagging
- Hindi Multitasking(POS tagging, gender and quantity representation(singular or plural)

[1]https://fasttext.cc/docs/en/unsupervised-tutorial.html

- Training a model on both Nepali and Hindi dataset together for POS tagging (Transfer learning)

## 8.5 Default setup

The default setup for these experiments is given below in table 8.3 which were derived from the paper put forth by Reimers et. al. [1] as they have conducted various experiments to find the optimal parameters to train a BLSTM-CNN-CRF model.

| Hyperparameter | Value |
|---|---|
| Dropout | 0.25 |
| Classifier | CRF |
| LSTM size | 100 |
| Epochs setup | 50 |
| Early stopping | 5 |
| Sentences per epoch | 3351(Hindi) & 3347(Nepali) |
| Dimension of Word embedding | 128 |
| Optimizer | Adam |
| Batch Size | 32 |
| Window size of CNN | 3 |
| Filter size of CNN | 30 |

**Table 8.3:** Default hyper-parameter setup

## 8.6 Results and Discussions

This section deals with the results obtained. The experiments were conducted using all three different embeddings types for POS tagging in Hindi, Nepali, and Multitask learning in Hindi. All these three embeddings were used to train the models with the varying following hyper-parameters:

- Default setup;
- Changing the dropout rate in default setup from 0.25 to 0.5;
- Changing the optimizer in default setup from Adam to AdaDelta.

This particular optimizer was chosen due to some research suggesting that it helps in improving the performance of the NLP models [76] [77]. Only combined Hindi and Nepali embedding were used for the Hindi-Nepali transfer learning POS tagging model. The models trained for in general 25-30 epochs as due to early stopping the training would stop if there is no improvement for a set number of epochs. For multitask learning only the accuracy of POS tagging is shown because the other tasks were used as auxiliary tasks to help in the improvement of the main POS tagging task. All the experiments were conducted three times.

### 8.6.1 Nepali

This section reports the results obtained by first training and then testing the BLSTM-CNN-CRF model on the POS-tagged Nepali dataset from CLE, Pakistan [66]. The results obtained from the various combination of embeddings and parameters with the Nepali Language are shown in figure 8.4. As can been seen in the table, with Nepali trained word embedding in the default setup, the accuracy of the model was in the range of 0.9133 to 0.9183. When the dropout was changed from 0.25 to 0.50 the accuracy seems to have improved to the range of 0.9158 to 0.9189. And finally, when the default ADAM optimizer was replaced by AdaDelta, the accuracy seems to go lower with a range of 0.9165 to 0.9174.
With Vector mapped word embeddings, in the default setup, the accuracy seems to be in the range of 0.8605 to 0.9092. When the dropout is increased to 0.5, then the accuracy seems to improve the range of 0.8904 to 0.8954. Finally, when the ADAM optimizer is replaced by AdaDelta, the accuracy seems to improve to the range of 0.9034 to 0.9063.
Finally, with Hindi-Nepali word embedding, in the default setup, the accuracy seems to be in the range of 0.9328 to 0.9335. When the dropout value is increased to 0.5, the accuracy seems to somehow improve to the range of 0.9324 to 0.9383. Finally, with a replacement of the default ADAM optimizer with AdaDelta, the accuracy seems to reduce to the range of 0.9298 to 0.9327.

| Embeddings | Default setup | Dropout:0.5 | AdaDelta Optimizer |
|---|---|---|---|
| Nepali | 0.9133 - 0.9183 | 0.9158 - 0.9189 | 0.9165 - 0.9174 |
| Vecmap Nepali | 0.8605 - 0.9092 | 0.8904 - 0.8954 | 0.9034 - 0.9063 |
| Hindi-Nepali | **0.9328 - 0.9335** | **0.9324 - 0.9383** | **0.9298 - 0.9327** |

**Table 8.4:** Result for Nepali POS tagging using different embeddings and parameters

When we look at all the different combinations of the experiment, it is pretty clear that in the case of embeddings, Hindi-Nepali combined embeddings seems to outperform normal Nepali only trained embedding and vector mapped Nepali embeddings. However, when we consider different setups, with Nepali and Hindi-Nepali combined embeddings, the setup with dropout value 0.5 seems to help in improving the performance of the model. Whereas for Vector mapped Nepali embeddings, setup with AdaDelta optimizer seems to outperform other setups.

### 8.6.2 Hindi

This section reports the results obtained by the first training and then testing the BLSTM-CNN-CRF model on the POS-tagged Hindi dataset from LDC with catalogue number LDC2010T24. The results obtained for various embeddings and parameters are shown in table 8.5. As it can be seen that with embeddings trained on Hindi texts only for the default setup the accuracy is in the range of 0.8599 to 0.8650. When the dropout is changed from 0.25 to 0.5 then the accuracy seems to reduce a little bit in the range of 0.8472 to 0.8628. And finally with the AdaDelta optimizer then it seems to increase a little bit to the range of 0.8593 to 0.8657.

| Embeddings | Default setup | Dropout:0.5 | AdaDelta Optimizer |
|---|---|---|---|
| Hindi | 0.8599 - 0.8650 | 0.8472 - 0.8628 | 0.8593 - 0.8657 |
| Vecmap Hindi | 0.8488 - 0.8528 | 0.8427 - 0.8482 | 0.8310 - 0.8460 |
| Hindi-Nepali | **0.8645 - 0.8702** | **0.8624 - 0.8638** | **0.8592 - 0.8682** |

**Table 8.5:** Result for Hindi POS tagging using different embeddings and parameters

With vector mapped Hindi embeddings, in the default setup, the accuracy seems to be in the range of 0.8488 to 0.8528. When the dropout is changed to 0.5 from default 0.25, the accuracy seems to lower down a bit to a range of 0.8427 to 0.8482. And finally with the AdaDelta optimizer the accuracy lowers down further to a range of 0.8310 to 0.8460. With the Hindi-Nepali joint embedding in the default setup, the accuracy seems to be in the range of 0.8645 to 0.8702. Then with changing the dropout to 0.5, the accuracy seems to lower down to the range of 0.8624 to 0.8638. And finally, with the AdaDelta optimizer, the accuracy seems to lower further down to a range of 0.8592 to 0.8682.

Hence, it can be seen from the above experiments that the Hindi-Nepali joint embedding helps the model to achieve the highest range of accuracy compared to other embeddings. The default setup and the setup with dropout 0.5 seems to help the model achieve an accuracy in a similar range.

### 8.6.3 MTL in Hindi

This section reports the results obtained by the first training and then testing the BLSTM-CNN-CRF model on the Hindi dataset from LDC with catalogue number LDC2010T24. The dataset is POS-tagged, gender-tagged and singular/plural tagged for each word. The task of POS-tagging was considered the main task in this thesis work and the other tasks were used as the auxiliary tasks. So the accuracy of the model on the other tasks is not reported.
The architecture of multi-task learning implemented here is of hard parameter sharing.

The accuracy on the task of POS tagging for multitask learning in the Hindi language is shown in figure 8.6. With Hindi only trained embeddings in the default setup, the accuracy is in the range of 0.7696 to 0.7708. When the dropout is further increased from 0.25 to 0.5 then the accuracy seems to lower a little bit to the range of 0.7661 to 0.7703. Then finally the ADAM optimizer is replaced by the AdaDelta optimizer, the accuracy further drops to the range of 0.7532 to 0.7612.
With vector mapped word embeddings in the default setup, the model seems to achieve a accuracy in the range of 0.7593 to 0.7621. When the dropout is changed to 0.5 then the accuracy drops a little bit to the range of 0.7567 to 0.7601. And finally, with the AdaDelta optimizer, the accuracy seems to further drop to the range of 0.7511 to 0.7592.

| Embeddings | Default setup | Dropout:0.5 | AdaDelta Optimizer |
|---|---|---|---|
| Hindi | 0.7696 - 0.7708 | 0.7661 - 0.7703 | 0.7532 - 0.7612 |
| Vecmap Hindi | 0.7593 - 0.7621 | 0.7567 - 0.7601 | 0.7511 - 0.7592 |
| Hindi-Nepali | **0.7866 - 0.7891** | **0.7843 - 0.7871** | **0.7634 - 0.7781** |

**Table 8.6:** Result for MTL Hindi POS tagging with various embeddings and parameters

With Hindi-Nepali joint embeddings in default mode, the model seems to achieve an accuracy in the range of 0.7866 to 0.7891. With an increase in dropout to 0.5 then the model seems to have an accuracy of 0.8624 to 0.8638. And finally, with AdaDelta Optimizer, the model has an accuracy in the range of 0.7634 to 0.7781.

Hence, it can be seen again that with Hindi-Nepali joint embeddings, the model achieves higher accuracy than with other embeddings, and with the default setup, the model achieves higher accuracy.

### 8.6.4 Transfer learning

The accuracy of POS tagging for the Nepali language with the transfer learning model using Hindi-Nepali embeddings can be seen in table 8.7. As it can be seen that in default setup the model achieves an accuracy in the range of 0.9303 to 0.9310. Then with the dropout being increased to 0.5, the model's accuracy does seem to improve. And finally, with replacing the ADAM optimizer with AdaDelta, the accuracy range seems to lower a bit to 0.9287 to 0.9311.

| Embeddings | Default setup | Dropout=0.5 | AdaDelta Optimizer |
|---|---|---|---|
| Hindi-Nepali | 0.9303 - 0.9310 | 0.9307 - 0.9319 | 0.9287 - 0.9311 |

**Table 8.7:** Result for Nepali TL POS tagging with different parameters

The accuracy of POS tagging for the Hindi language with the transfer learning model using Hindi-Nepali embeddings can be seen in table 8.5. As can be seen with the default setup the model achieves an accuracy in the range of 0.7625 to 0.7652. Then with the dropout 0.5, the accuracy seems to improve to the range of 0.7725 to 0.7741, and finally, with AdaDelta optimizer, the model achieves an accuracy in the range of 0.7627 to 0.7629.

| Embeddings | Default setup | Dropout=0.5 | AdaDelta Optimizer |
|---|---|---|---|
| Hindi-Nepali | 0.7625 - 0.7652 | 0.7725 - 0.7741 | 0.7627 - 0.7629 |

**Table 8.8:** Result for Hindi TL POS tagging with different parameters

For both the languages, it seems that the default setup and setup with dropout value of 0.5 provides accuracy in similar range. For both the languages, it is observed that setup with AdaDelta optimizer doesn't help the optimizer in improving its performance.

### 8.6.5 Summary

The results obtained in all the experiments in the previous subsection can be summarized as:

- From the experiments conducted it has been observed that, Hindi-Nepali joint embeddings help the model perform best in all the experiments. The possible reason for this could be that when the word embeddings are trained on both the languages together, the vector representation for a particularly common word is based on its similarity with words in both languages and not just on one language. This knowledge seems to help the model in its task of POS tagging.

- In the present context of experiments, the unsupervised vector mapping between monolingual embeddings of Hindi and Nepali language doesn't seem to improve the performance of the model, on the contrary, it lowers the accuracy compared to the normal individual language embeddings. The possible reason for this could be the fact that when the embeddings are trained monolingually, the vector for a word is created on the basis of the words in its surrounding in that particular language. Hence when the embeddings are mapped between two languages the vector space orientation is shifted which seems to be detrimental to the performance of the model.

- As it has been observed in the above experiments, default setup which is inspired from the work of Reimers et. al [1] seems to achieve the highest accuracy in case of individual POS tagging in Hindi, Nepali, and Multitasking in Hindi whereas the setup with dropout 0.5 and the defalult setup seems to achieve accuracy in similar range with the transfer learning model.

- It has been observed from the experiments that the accuracy achieved by the model on Hindi dataset is lower than the Nepali counterpart. The possible reason for this could be the fact that sentences in Hindi dataset are relatievely longer and more complicated than Nepali dataset. However, there is a need for further indepth exploration into other reasons.

- In the above experiments. it has been observed that transfer learning or training the model jointly on similar tasks in two domains doesn't seem to help the model's performance as much. In the case of Nepali, the accuracy seems to lower just by little but in the case of Hindi, it has further lowered notably. One of the possible reasons could be that as the performance of the model was not so good on Hindi dataset, it led to the overall poor performance of the model. And other possible reason could be that even if the languages are similar, this particular deep learning architecture could not exploit its similarity. Hence, the other transfer learning architectures like sequential transfer learning could also be explored.

- In this experimental context, multitask learning in the Hindi language for the task of POS tagging with auxiliary tasks of gender tagging and singular/plural tagging doesn't seem to help the model at all. On the contrary, the performance of the model on the primary task of the POS tagging has lowered. The possible reason for this could be that the knowledge from the auxiliary tasks like gender tagging and singularity/plurality tagging were not helpful for the main task and hence led to reduction in the performance of the model.

- It has been observed in the above experiments that the AdaDelta optimizer doesn't seem to help in the performance of the model as in all the results the accuracy is relatively lower compared to Adam optimizer. The possible reason for it could be that it's convergence rate is dependent on initial learning rate and it seems that chosen learning rate is not suitable for the optimizer.

# 9 Conclusion and Future Work

This research work was primarily focused on exploring the plausibility of using transfer learning in deep neural network architectures (BLSTM-CNN-CRF in this case) between rarely annotated languages like Hindi and Nepali language to explore if it can be helpful in improving the performance of the model in solving similar tasks (POS tagging in this case) in both the languages. Various experiments were conducted with different embeddings and with different setups and respective results were discussed. The author would like to conclude this research work by trying to answering the research questions which led to this work.

- ***Can embeddings trained in both Hindi and Nepali languages help a deep learning model perform better on common tasks in both the languages?***
  A trend has been observed in all the experiments in sections 8.6.1 to 8.6.3 that Nepali-Hindi joint embedding did help the model in achieving higher accuracy than the normal monolingual embeddings and the cross-lingual vector mapped embedding. Hence it can be concluded that using a bilingual embedding of similar languages helps the deep learning model to achieve better performance.

- ***Can transfer learning between Hindi and Nepali language help a model perform better on the task of POS tagging in both languages?***
  It can be observed from the the experiments conducted that the transfer learning or jointly training the model for the task of POS tagging in both the languages has not been so helpful for the model. It seems to lower the performance of the model compared to POS tagging with just a single language model. Hence, in this context it could be concluded that transfer learning was not helpful for the task of POS-tagging.

- ***Can multitask learning in Hindi with the main task of POS tagging and auxiliary tasks of gender tagging and singularity/plurality tagging, help in the improvement of the performance of the model in the task of POS tagging?***
  As it can be observed in the results that multitasking doesn't seem to help in the task of POS tagging in the Hindi language with auxiliary tasks of gender tagging and singularity/plurality tagging. On the contrary, it seems to lower the performance of the model on the task of POS tagging. Hence, from the above experiments, it can be concluded that multitasking was not helpful for the task of POS tagging with auxiliary tasks of gender tagging and singularity/plurality tagging.

- ***Can unsupervised vector mapping the word embeddings of Hindi and Nepali cross-lingually help in improving the performance of the model?*** The results depict that it is not at all helpful to used cross-lingually mapped embedding for the task of POS tagging in both the languages as it seems to lower the performance of the model compared to normal embeddings. Hence, it can be safely concluded that for the task of POS tagging in Hindi and Nepali language, the vector mapping doesn't seem to help.

This thesis work explores just the BLSTM-CNN-CRF model for the transfer learning between Hindi and Nepali language and Multitasking in the Hindi language. However, the above experiments could also be conducted on various other models like LSTM-CRF+ELMo+BERT, Bi-LSTM-CRF+ElMo, and various other neural network architectures. As the Hindi-Nepali combined trained embedding seems to provide a better performance, statistical tests could be further conducted to make quantitative conclusions. And the embeddings could also be further used with other architectures. From the experiments, it seems the unsupervised cross-lingual vector mapped embeddings instead of improving the performance, it degrades the performance of the model but there are supervised cross-lingual embeddings and semi-supervised cross-lingual embeddings which could be further explored for the task. It could also be seen that the ADAM optimizer seems to provide better accuracy than the AdaDelta optimizer. As the AdaDelta optimizer's performance depends highly on the initial learning rate, further experiments could be conducted with varying learning rates. This model could also be used to explore transfer learning between other languages with similarities. The performance of the model on the Hindi langauge dataset was not satisfactory and further reasons behind it could be explored. The multitask learning in the Hindi language doesn't seem to help in improving the performance of the model for POS tagging with auxiliary tasks of Gender tagging and singularity/plurality tagging, hence other auxiliary tasks. Transfer learning between Nepali and Hindi doesn't seem to help the task of POS tagging. The reason behind this needs to be explored in more depth. As it is just one architecture of transfer learning and hence other architectures like sequential transfer learning could also be explored for the task.

# Statutory Declaration

I herewith declare that I have composed the present thesis myself and without use of any other than the cited sources and aids. Sentences or parts of sentences quoted literally are marked as such; other references with regard to the statement and scope are indicated by full details of the publications concerned. The thesis in the same or similar form has not been submitted to any examination body and has not been published. This thesis was not yet, even in part, used in another examination or as a course performance.

Rostock, 20.07.2020

---

Dipendra Yadav